\PassOptionsToPackage{table}{xcolor} %
\documentclass[sigconf]{acmart}
\usepackage[table]{xcolor}
\usepackage{amsmath}
\usepackage{tabularx}
\usepackage[utf8]{inputenc}
\usepackage{pifont}
\usepackage{tabularx}
\usepackage{booktabs}
\usepackage{xspace}
\usepackage{graphicx}
\usepackage{tikz}
\usetikzlibrary{fit,positioning,calc,shapes,backgrounds}
\usepackage{pgfplots}
\pgfplotsset{compat=1.14}
\usepgfplotslibrary{fillbetween}
\usepackage{pgfplotstable} 
\usepackage{filecontents}

\usepackage{epigraph}
\usepackage[caption=false]{subfig}
\usepackage{caption}
\usepackage{enumitem}
\usepackage[framemethod=tikz]{mdframed}
\usepackage{todonotes}
\usepackage{hyphenat} %
\usepackage{xfrac}
\usepackage{multirow}
\newcommand{\specialcell}[2][c]{%
    \begin{tabular}[#1]{@{}c@{}}#2\end{tabular}}
\usepackage{balance}
\usepackage[linesnumbered,lined,boxed]{algorithm2e}
\usepgfplotslibrary{groupplots}
\usepackage{algorithmic}
\usetikzlibrary{pgfplots.groupplots}
\usepackage{xcolor}
\definecolor{green1}{RGB}{0,153,0}
\definecolor{red1}{RGB}{204,0,0}
\definecolor{blue1}{RGB}{0,0,153}
\usepackage{color}
\usepackage{pgfplotstable}
\usetikzlibrary{arrows,petri,topaths,backgrounds,snakes,patterns,positioning}
\usepackage{tkz-berge}

    \usetikzlibrary{
        pgfplots.colorbrewer,
    }
    \pgfplotsset{
        cycle list/.define={my marks}{
            every mark/.append style={solid,fill=\pgfkeysvalueof{/pgfplots/mark list fill}},mark=*\\
            every mark/.append style={solid,fill=\pgfkeysvalueof{/pgfplots/mark list fill}},mark=square*\\
            every mark/.append style={solid,fill=\pgfkeysvalueof{/pgfplots/mark list fill}},mark=triangle*\\
            every mark/.append style={solid,fill=\pgfkeysvalueof{/pgfplots/mark list fill}},mark=diamond*\\
        },
    }

\tikzstyle{stateTransition}=[-stealth, thick]
\tikzstyle{inputNode}=[draw,circle,minimum size=10pt,inner sep=0pt]

\newtheorem{definition}{Definition}

\makeatletter
\g@addto@macro\normalsize{%
  \setlength\abovedisplayskip{1pt}
  \setlength\belowdisplayskip{1pt}
  \setlength\abovedisplayshortskip{1pt}
  \setlength\belowdisplayshortskip{1pt}
}
\makeatother

\copyrightyear{2021}
\acmYear{2021}
\acmConference[WWW '21]{Proceedings of the Web Conference 2021}{April 19--23, 2021}{Ljubljana, Slovenia}
\acmBooktitle{Proceedings of the Web Conference 2021 (WWW '21), April 19--23, 2021, Ljubljana, Slovenia}
\acmPrice{}
\acmDOI{10.1145/3442381.3449882}
\acmISBN{978-1-4503-8312-7/21/04}

\newcommand{\ourmethod}{Pathfinder Discovery Networks}

\title{Pathfinder Discovery Networks for Neural Message Passing}

\begin{document}

\author{Benedek Rozemberczki}
\authornote{Work done while interning at Google.}
\affiliation{
 \institution{The University of Edinburgh}
}
\email{benedek.rozemberczki@ed.ac.uk}
\author{Peter Englert}
\authornote{Now at Amazon Japan (englertp@amazon.co.jp)}
\affiliation{
  \institution{Google Research}
}
\author{Amol Kapoor}
\affiliation{
  \institution{Google Research}
}
\email{ajkapoor@google.com}

\author{Martin Blais}
\affiliation{%
  \institution{Google Research}
}
\email{blais@google.com}

\author{Bryan Perozzi}
\affiliation{%
  \institution{Google Research}
}
\email{bperozzi@acm.org}

\begin{abstract}

In this work we propose \textit{\ourmethod} (PDNs), a method for jointly learning a message passing graph over a multiplex network with a downstream semi-supervised model. PDNs inductively learn an aggregated weight for each edge, optimized to produce the best outcome for the downstream learning task. PDNs are a generalization of attention mechanisms on graphs which allow flexible construction of similarity functions between nodes.  They also support edge convolutions and cheap multiscale mixing layers. We show that PDNs overcome weaknesses of existing methods for graph attention (e.g. Graph Attention Networks), such as the diminishing weight problem.

Our experimental results demonstrate competitive predictive performance on academic node classification tasks. Additional results from a challenging suite of node classification experiments show how PDNs can learn a wider class of functions than existing baselines. We analyze the relative computational complexity of PDNs, and show that PDN runtime is not considerably higher than static-graph models. Finally, we discuss how PDNs can be used to construct an easily interpretable attention mechanism that allows users to understand information propagation in the graph.

\end{abstract} 
\maketitle

\section{Introduction}


Recently, there has been a surge of interest in applying neural networks to graph data. The last few years have seen the development of a wide variety of approaches, ranging from graph embedding \citep{deepwalk, node2vec, diff2vec, postuavaru2020instantembedding}, to graph convolutional networks \cite{kipf2017semi,graphsage_nips17}, to message passing neural networks \citep{gilmer2017neural}.
Though powerful, many of these approaches have a serious limitation: they assume that the underlying graph is static, provided as an immutable input parameter where edges between node-pairs have only a single weight. However, in many real world applications, there is rarely one `correct' graph -- instead, the best task performance comes from combining many different types of relationships  \cite{grale}.  For example, in a video classification task, the best graph to use might consider several different types of similarity between videos (e.g.\ both image similarity and audio similarity). In practice, we believe that it is a mistake to separate graph construction from the learning task at hand; rather, the optimal graph must come from deep consideration of the problem being solved.
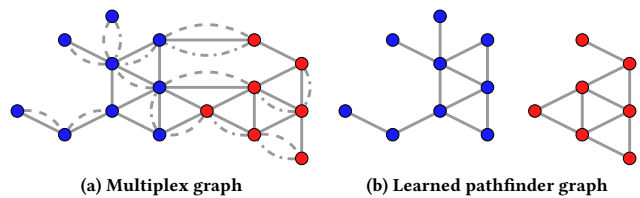
\begin{figure}[h!]
	\centering
	\subfloat[Multiplex graph\label{subfig:multiples}]{\begin{tikzpicture}[scale=0.315,transform shape]
		\tikzstyle{VertexStyle}=[minimum size = 15pt,inner sep=0pt]
        \tikzset{VertexStyle/.append style={rectangle}}		
		\Vertex[L=$$,x=0,y=0]{0}
		\Vertex[L=$$,x=2,y=-1]{1}
		\Vertex[L=$$,x=4,y=0]{2}
		\Vertex[L=$$,x=4,y=2]{3}
		\Vertex[L=$$,x=2,y=3]{4}
		\Vertex[L=$$,x=6,y=-1]{5}
		\Vertex[L=$$,x=6,y=1]{6}
		\Vertex[L=$$,x=6,y=3]{7}
		\Vertex[L=$$,x=4,y=4]{14}	
		
        \tikzset{VertexStyle/.append style={circle}}
		\Vertex[L=$$,x=8,y=0]{8}
		\Vertex[L=$$,x=10,y=-1]{9}
		\Vertex[L=$$,x=10,y=1]{10}
		\Vertex[L=$$,x=12,y=0]{11}
		\Vertex[L=$$,x=12,y=2]{12}	
		\Vertex[L=$$,x=10,y=3]{13}	
		\Vertex[L=$$,x=12,y=-2]{15}		
		
		\AddVertexColor{blue!90, draw=black}{0,1,2,3,4,5,6,7, 14}
		\AddVertexColor{red!90, draw=black}{8, 9, 10, 11, 12, 13, 15}
		
		\tikzstyle{EdgeStyle}=[line width=1.1pt, opacity = 0.4]

		\Edge[label=](5)(8)
		\Edge[label=](6)(8)
		\Edge[label=](6)(10)
		\Edge[label=](7)(13)
		
		\Edge[label=](8)(9)	
				
		\Edge[label=](8)(10)
		\Edge[label=](9)(10)
		\Edge[label=](11)(9)		
		\Edge[label=](11)(10)
		\Edge[label=](12)(10)		
		\Edge[label=](11)(12)
		\Edge[label=](13)(12)
		\Edge[label=](15)(11)	
		\Edge[label=](0)(1)
		\Edge[label=](1)(2)
		\Edge[label=](3)(2)
		\Edge[label=](3)(4)
		\Edge[label=](2)(5)
		\Edge[label=](2)(6)
		\Edge[label=](3)(6)
		\Edge[label=](3)(7)
		\Edge[label=](5)(6)
		\Edge[label=](6)(7)
		\tikzstyle{EdgeStyle}=[line width=1.1pt, opacity = 0.4, dashed, bend left]
		\Edge[label=](0)(1)
		\Edge[label=](1)(2)
		\Edge[label=](3)(4)
		\Edge[label=](6)(3)
		\Edge[label=](8)(5)
		\Edge[label=](3)(14)
		\Edge[label=](9)(15)
		\Edge[label=](6)(10)
		\Edge[label=](7)(13)		
		\tikzstyle{EdgeStyle}=[line width=1.1pt, opacity = 0.4, dash dot, bend left]
		\Edge[label=](5)(6)
		\Edge[label=](7)(3)
		\Edge[label=](9)(8)
		\Edge[label=](12)(11)
		\Edge[label=](11)(10)
		\Edge[label=](14)(3)
		\Edge[label=](15)(9)	
		\Edge[label=](13)(7)
		\end{tikzpicture}}
	\hspace{5pt}
	\subfloat[Learned pathfinder graph\label{subfig:learned_graph}]{
\begin{tikzpicture}[scale=0.315,transform shape]
		\tikzstyle{VertexStyle}=[minimum size = 15pt,inner sep=0pt, shape = circle]
		\Vertex[L=$$,x=0,y=0]{0}
		\Vertex[L=$$,x=2,y=-1]{1}
		\Vertex[L=$$,x=4,y=0]{2}
		\Vertex[L=$$,x=4,y=2]{3}
		\Vertex[L=$$,x=2,y=3]{4}
		\Vertex[L=$$,x=6,y=-1]{5}
		\Vertex[L=$$,x=6,y=1]{6}
		\Vertex[L=$$,x=6,y=3]{7}
		\Vertex[L=$$,x=8,y=0]{8}
		\Vertex[L=$$,x=10,y=-1]{9}
		\Vertex[L=$$,x=10,y=1]{10}
		\Vertex[L=$$,x=12,y=0]{11}
		\Vertex[L=$$,x=12,y=2]{12}	
		\Vertex[L=$$,x=10,y=3]{13}	
		\Vertex[L=$$,x=4,y=4]{14}				
		\Vertex[L=$$,x=12,y=-2]{15}	

		\AddVertexColor{blue!90, draw=black}{0,1,2,3,4,5,6,7, 14}
		\AddVertexColor{red!90, draw=black}{8, 9, 10, 11, 12, 13, 15}
		
		\tikzstyle{EdgeStyle}=[line width=1.1pt, opacity = 0.4]
		\tikzstyle{LabelStyle}=[fill=white]

		\Edge[label=](8)(9)		
		\Edge[label=](8)(10)
		\Edge[label=](9)(10)
		\Edge[label=](11)(9)		
		\Edge[label=](11)(10)
		\Edge[label=](12)(10)		
		\Edge[label=](11)(12)
		\Edge[label=](13)(12)
		\Edge[label=](15)(11)	
		\Edge[label=](15)(9)	
		\Edge[label=](0)(1)
		\Edge[label=](1)(2)
		\Edge[label=](3)(2)
		\Edge[label=](3)(4)
		\Edge[label=](2)(5)
		\Edge[label=](2)(6)
		\Edge[label=](3)(6)
		\Edge[label=](3)(7)
		\Edge[label=](5)(6)
		\Edge[label=](6)(7)
		\Edge[label=](14)(3)
		\end{tikzpicture}
		
	}	

	\caption{Pathfinder discovery networks take multiple sets of weighted edges and learn a graph specifically suited to a downstream predictive task. In our example multiplex graph (Fig.\ \ref{subfig:multiples}) we have three types of edges and a two types of nodes. The pathfinder discovery network would perform node classification and output a learned pathfinder graph where inter-class edges are forgotten (Fig.\  \ref{subfig:learned_graph}).}\label{fig:spotlight}
\end{figure}
Some methods have attempted to relax these limitations.
For example, \textit{Graph Attention Networks} (GATs) \cite{gat_iclr18}, attempt to re-weight each edge in the graph.
However, GAT models are prone to overfitting, and due to their over-parameterization have difficulties being trained on large real world datasets. 
Further, they suffer from a diminishing weight problem, which drives learned edge weights towards zero as the degree of a node increases.
Finally (and perhaps most importantly) GAT-style attention constrains edge reweighting to be a single aggregation of the node features, which prevents GATs from learning complex difference operators between edges in varying neighborhood structures.

A separate body of work has proposed specific models for heterogeneous data \citep{zhang2018deep, yao2019graph}. However these models are often highly specific to specific kinds of data inputs (e.g. a model might support video and relational data, but not geospatial data), and are therefore difficult to integrate with new advances in modeling. More to the point, though heterogeneous approaches do incorporate a wide variety of data types, they still treat graph construction as a fundamentally isolated problem from graph learning. As a result, heterogeneous approaches will struggle for the same reasons mentioned above.


In this work, we answer the question: \textit{``How can we learn to construct the optimal graph for solving any given learning problem?''}. 
For inspiration, we look back to \emph{pathfinder networks} \cite{schvaneveldt1989network}, a graph construction measure from the psychology literature.
In a pathfinder network, multiple kinds of proximity judgements (e.g.\ relatedness scores from humans) are considered simultaneously in order to determine edges between node-pairs.
A discrete algorithm finds the graph which best preserves some property (e.g.\ shortest paths) of the input proximities.
While traditional pathfinder networks are useful for tasks such as aggregating subjective information, they are unfortunately unsuitable for use in most graph learning tasks.
More recently, Halcrow et.\ al \cite{grale} describe a similar problem of graph construction from multiple proximities which occurs in a wide variety of industrial applications.
Their solution, Grale, uses a model to precompute a fused similarity network for graph learning (one edge at a time).
While this system has many advantages (scalability, allows use of different kinds of relationships, etc) and has been used in a wide variety of applications at Google, it is not able to learn a graph jointly with a downstream task.
Here, we go one step further, and present \emph{\ourmethod} (PDNs), a framework for learning a network over a set of entities with diverse similarity scores jointly with a graph neural network task. 

Our main contribution is the design and validation of the \textit{pathfinder layer} - a differentiable neural network layer which is able to combine multiple sources of proximity information defined over a set of nodes to form a single weighted graph. The pathfinder layer uses a feed forward neural network to learn the edge weights while the sparsity of the underlying weighted multiplex graph is unchanged (see Figure \ref{fig:spotlight}). This layer feeds directly into a downstream GNN model that is set up to learn arbitrary tasks -- in this paper, semi-supervised classification. Gradients from the supervised classification task propagate down to the edge weights, allowing PDNs to create a graph that is optimized for the classification task at hand. Our model learns this graph in an inductive manner, allowing transfer of learned graph aggregation from one graph to another.

We demonstrate the flexibility of our framework by showing that a number models can be seen as special cases of our general framework. First, we show how edge convolutional models can be formulated with our modeling framework. Second, we establish that one can define models that perform cheap multi-scale mixing with the pathfinder layer. 

Our empirical analysis focuses on node classification tasks, feature importance measurements, and runtime comparisons. We use synthetic experiments to demonstrate a class of learnable tasks where PDNs significantly outperform current state-of-the-art methods thanks to the unique ability to learn arbitrary functions over multiple proximity inputs. We then switch to real world node classification problems, and demonstrate that PDN has a 0.8\%-3.5\% predictive performance advantage over the most competitive existing graph neural network models in terms of accuracy. We analyze the runtime of PDNs and demonstrate that the pathfinder layer increases the training runtime by a constant multiplier. Finally, we describe how the weights of the pathfinder layer can be seen as attention over the input graphs and edge features, and add interpretability to the underlying information propagation.
The key contributions of our work are as follows:
\begin{enumerate}
    \item We propose a flexible framework to learn a single graph for message passing from multiple graphs jointly with any graph convolution layer. 
    \item We showcase how this framework can be used to define edge convolution neural network models where the message passing graph is learned from node features.
    \item We define models with cheap multi-scale mixing where the adjacency matrix of the message passing graph is a linear combination of adjacency matrix powers.
    \item We empirically demonstrate that our models have competitive results on a range of node classification tasks, have decent runtimes on small-scale graphs, and have explainable weights in case of the simple models.
\end{enumerate}
The source code of Pathfinder Discovery Networks is available at \url{https://github.com/benedekrozemberczki/PDN}.

\section{Preliminaries}\label{sec:related_work}
We begin by summarizing the notation used in our work and reviewing the related concepts of multiplexity, simplified spectral graph convolutions, and graph attention. We frame our model as a general building block that can be applied to a wide variety of graph neural network designs.

\textbf{Notation.} We assume that we have a set of vertices $V$, and $D$ graphs defined on these vertices described by $\mathcal{G}_1,\dots,\mathcal{G}_D$ with respective edge sets $E_1,\dots,E_D$. These graphs can be represented as $|V| \times |V|$ adjacency matrices which are respectively denoted by $\widetilde{\textbf{A}}_1,  \dots , \widetilde{\textbf{A}}_D$. We assume that nodes have generic vertex features. For the whole set $V$, these features are described by a feature matrix $\textbf{X} \in \mathbb{R}^{|V| \times F}$, where $F$ is the number of features. In addition, for each node we have a target that we want to predict. For the whole set $V$, the targets are defined as a $|V| \times C$ binary matrix, where $C$ is the number of node classes. Our goal is to predict the target class matrix using the graphs and the node features.

\textbf{Multiplex graphs and learning.} This problem setup can be framed as node classification with a \textit{weighted multiplex graph} \citep{menichetti2014weighted} which has no inter-layer edges (Figure \ref{subfig:multiples}). Unlike heterogeneous graphs, multiplex graphs operate on a single node type. Current approaches to learning from multiplex graphs only generalize neighbourhood based embeddings to accommodate a multiplex setting \cite{matsuno2018mell,zhang2018scalable}. In these approaches, a separate node embedding is learned for each graph (or each layer in the multiplex graph); these embeddings are concatenated to form the node representations. These approaches have two limitations.  First, the node embeddings are transductive so the models do not generalize from one graph to another. Second, these approaches are expensive, as node level embeddings must be calculated for each graph separately.

\textbf{Graph convolutions.} The traditional setting of spectral graph convolutional networks \cite{kipf2017semi} has a single graph $\mathcal{G}$ and a corresponding adjacency matrix $\textbf{A}$. In the forward pass of the spectral model the degree normalized adjacency matrix $ \textbf{D}^{-1/2} \textbf{A} \textbf{D}^{-1/2} $ is used to propagate the hidden node representations. These hidden representations are obtained by multiplying the feature matrix $\textbf{X}$ by a trainable $F \times d$ weight matrix $\textbf{W}$. Finally, the aggregated representations are transformed by $\sigma(\cdot)$ an elementwise non-linearity just as in Equation \eqref{eq:basic_gcn_layer} which describes the whole forward pass.
\begin{align}
\mathbf{Z} &=\sigma( \textbf{D}^{-1/2} \textbf{A} \textbf{D}^{-1/2} \textbf{X} \textbf{W})\label{eq:basic_gcn_layer}
\end{align}
As defined, the spectral graph convolutional model cannot accommodate the presence of multiple graphs -- one either has to come up with a pre-defined edge weight aggregation function or use only one of the graphs from $\mathcal{G}_1,\dots,\mathcal{G}_D$ as the message passing graph. PDNs are directly motivated by this fundamental weakness. We note that a range of graph neural network architectures use the spectral graph convolution as a building block \cite{chami2020machine}. We believe that overcoming the single graph limitation could therefore lead to significant improvements in all other related models.

\textbf{Graph attention networks.} Graph attention networks \citep{gat_iclr18} learn the edge weights used for message passing using the features of nodes at the edge endpoints. Node features are transformed by a learnable parameter matrix and concatenated together for each edge. The node-pair representations are multiplied by an attention vector, and the weight of each edge is decided by a softmax unit defined over the neighbors of the source node. Although the GAT model can learn multiple edge weights with multiple attention heads, this is not a straightforward comparison to learning multiplex graphs. In \citep{gat_iclr18}, each attention head is trained on the \textit{same node features}, which precludes incorporating unique features from multiple sources. 
Furthermore, GAT cannot fully leverage the power of multiple attention heads, because the final edge weight is a simple average over the individual heads (and not a learned function). The GAT model is therefore unable to learn an expressive range of functions over multiple edge sets, which severely limits its applicability to multiplex graph problems.

\textbf{Pathfinder discovery networks as a building block.} The pathfinder layer introduced in our work is sufficiently general to serve the message passing matrix for a wide range of general graph neural network models defined on multiplex graphs. A pathfinder discovery network can output a graph defined over a set of nodes with a single edge weight for each edge; as a result, PDNs are easily applied to any neural models that use the graph directly (including \textit{Spectral Graph Convolutions} \citep{kipf2017semi}, \textit{Graph Sampling and Aggregation} \citep{graphsage_nips17, ying2018graph}, \textit{Multi-Scale Graph Convolutions} \citep{ mixhop_icml19,abu2019ngcn}, \textit{Clustered Graph Convolutions} \citep{clustergcn_kdd19}, \textit{Personalized Propagation of Neural Predictions} \citep{ppnp_iclr19,bojchevski2020pprgo} and \textit{Simplified Graph Convolutions} \citep{sgc_icml19}). We further note that pathfinder layers can be used as a building block for learning tasks beyond node prediction. For example, using pathfinder layers, an appropriate graph convolutional layer, and graph level pooling such as \textit{Sort Pooling} or \textit{Diff Pooling} \cite{dgcnn_aaai18, diffpool_nips18}, one can easily define models which characterize or classify whole graphs. 

\section{Message passing on learned graphs}\label{sec:base_model}
Our model jointly learns a single graph from a set of similarity graphs, and a graph neural network which uses the adjacency matrix of this learned graph as a propagation matrix. The adjacency matrices describing the input graphs themselves can be learned or pre-computed. An exemplar graph could be a set of $k$-nearest neighbor graphs of pairwise similarities calculated from multimodal datapoints, with separate graphs for images, sound, and text. Another potential example could be the use of normalized adjacency matrix powers as measures of pairwise similarity between nodes. In general, we can include as much information as possible with the expectation that the PDN will find the optimal graph structure based on consideration of all feature correlations.

\subsection{Pathfinder Learning Layers}
Here we detail the design of \textit{Pathfinder Learning Layers} -- neural network architectures for combining different kinds of proximity data together.  
We begin with the consideration of a simple model for combining proximity information, and extend it to support modeling complex relationships over multiplex data.
\begin{definition}
Pathfinder Neuron.  \emph{A pathfinder neuron (Fig \ref{fig:neuron}) takes weighted adjacency matrices} $\widetilde{\textbf{A}}_1, \dots, \widetilde{\textbf{A}}_D$  \emph{as input and combines them into a single $|V| \times |V|$ learned graph $\widetilde{\textbf{G}}$ as its output. 
It uses trainable weights to learn the relative importance of each kind of similarity information, as follows:}
\begin{align}
\widetilde{\textbf{G}} &= \sigma \left(\sum_{i=1}^{D} \beta_{i}\cdot \widetilde{\textbf{A}}_i\right).\label{eq:neuron}
\end{align}
\end{definition}
\noindent The elementwise function $\sigma(\cdot)$ is a non-linearity and $\beta_i$ is a trainable weight specific to the $i^{th}$ input adjacency matrix (see Figure \ref{fig:neuron}). We assume that there is no bias term present, which implies: (i) calculating $\widetilde{\textbf{G}}$ can be done entirely with sparse linear algebra operations; (ii) the output graph from one pathfinder neuron can be used as an input to other pathfinder neurons. 
Further, we note that multiple pathfinder neurons can take in the same inputs and learn different weights, akin to the GAT multi-attention-head.

\begin{definition}
Pathfinder Layer. \emph{A pathfinder layer uses multiple pathfinder neurons as building blocks for a more complex neural model.
The $l^{th}$ \textit{pathfinder layer} with $q$ neurons using $p$ input graphs can be written by:}
\begin{align}
\widetilde{\mathbf{G}}_{l+1,1};\dots; \widetilde{\mathbf{G}}_{l+1,q} =f^{l}\left(\widetilde{\mathbf{G}}_{l,1};\dots; \widetilde{\mathbf{G}}_{l,p}\right).\label{eq:grale_layer}
\end{align}
\end{definition}
\noindent Each $\widetilde{\mathbf{G}}$ is output by a single pathfinder neuron. The number of parameters in the neuron depends on the number of pathfinder graphs in the previous layer. 

\begin{definition}
Pathfinder Graph.  \emph{The final output of a pathfinder neuron, or a series of pathfinder layers in a pathfinder discovery network is the \emph{pathfinder graph}, denoted by $\hat{\textbf{G}}$.}
\end{definition}

The pathfinder graph can be used for message passing in an arbitrary downstream graph convolutional model (see Figure \ref{fig:deep_model}). If the edge sets of input graphs sufficiently overlap and the original graphs are sparse, we expect $\hat{\mathbf{G}}$ to be sparse.   
\subsection{Pathfinder Discovery Networks}
The pathfinder graph described above can be used as an input for an arbitrary downstream graph neural network. As a motivating example, consider the spectral graph convolutional network defined by Equation \ref{eq:basic_gcn_layer}. We can augment this equation by replacing $\textbf{A}$ with the final pathfinder graph:
\begin{align}
\mathbf{Z} &=\hat{\sigma}( \textbf{D}^{-1/2}_{\hat{\textbf{G}}} \hat{\textbf{G}} \textbf{D}^{-1/2}_{\hat{\textbf{G}}} \textbf{X} \textbf{W})\label{eq:gcn_layer}
\end{align}

\begin{figure}[h!]
\subfloat[Architecture of a single pathfinder neuron\label{fig:neuron}]{
    
\begin{tikzpicture}[scale=1.0,transform shape]

	\node[draw,circle,minimum size=25pt,inner sep=0pt] (x) at (9,-5) {$\sigma(\cdot)$};
	
	\draw[stateTransition] (x) -- (12,-5) node [midway,above] {$\sigma\left(\sum\limits_{i=1}^{D}{\beta_i \cdot \tilde{\textbf{A}}_i}\right)$};
	\node[inputNode] (x0) at (6, -3.5) {$\widetilde{\textbf{A}}_1$};
	\node[inputNode] (x1) at (6, -4.25) {$\widetilde{\textbf{A}}_2$};
	\node[inputNode] (x2) at (6, -5.0) {$\widetilde{\textbf{A}}_3$};
	\node[inputNode] (x3) at (6, -5.75) {$\widetilde{\textbf{A}}_4$};
	\node[inputNode] (xn) at (6, -6.5) {$\widetilde{\textbf{A}}_D$};
	\node[draw,circle,minimum size=17pt,inner sep=0pt] (tt) at (12.3,-5) {$\widetilde{\textbf{G}}$};

	\draw[stateTransition] (x0) to[out=0,in=120] node [midway, sloped, above] {$\beta_1$} (x);
	\draw[stateTransition] (x1) to[out=0,in=150] node [midway, sloped, above] {$\beta_2$} (x);
	\draw[stateTransition] (x2) to[out=0,in=180] node [midway, sloped, above] {$\beta_3$} (x);
	\draw[stateTransition] (x3) to[out=0,in=210] node [midway, sloped, above] {$\beta_4$} (x);
	\draw[stateTransition] (xn) to[out=0,in=240] node [midway, sloped, above] {$\beta_D$} (x);

\end{tikzpicture}
}%
    \qquad
\subfloat[A PDN consisting of one hidden pathfinder layer and a GCN \label{fig:deep_model}]{
 
\vspace{-3mm}
    
\begin{tikzpicture}[scale=1.0,transform shape]

	\node[draw,circle,minimum size=17pt,inner sep=0pt] (xy) at (10.8,-1.75) {$\textbf{X}$};
	\node[draw,circle,minimum size=17pt,inner sep=0pt] (z) at (12.5,0) {$\textbf{Z}$};	

	\node[inputNode, thick] (i1) at (6, 0.75) {};
	\node[inputNode, thick] (i2) at (6, 0) {};
	\node[inputNode, thick] (i3) at (6, -0.75) {};
	
	\node[inputNode, thick] (h1) at (8, 1.5) {};
	\node[inputNode, thick] (h2) at (8, 0.75) {};
	\node[inputNode, thick] (h3) at (8, 0) {};
	\node[inputNode, thick] (h4) at (8, -0.75) {};
	\node[inputNode, thick] (h5) at (8, -1.5) {};
	
	\node[inputNode, thick] (o1) at (9, 0.0) {};
	\node[draw,circle,minimum size=25pt,inner sep=0pt] (p) at (10.8,0) {$\hat{\sigma}(\cdot)$};
	
	\draw[stateTransition] (5, 0.75) -- node[above] {$\widetilde{\textbf{A}}_1$} (i1);
	\draw[stateTransition] (5, 0) -- node[above] {$\vdots$} (i2);
	\draw[stateTransition] (5, -0.75) -- node[above] {$\widetilde{\textbf{A}}_D$} (i3);
	
	\draw[stateTransition] (i1) -- (h1);
	\draw[stateTransition] (i1) -- (h2);
	\draw[stateTransition] (i1) -- (h3);
	\draw[stateTransition] (i1) -- (h4);
	\draw[stateTransition] (i1) -- (h5);
	\draw[stateTransition] (i2) -- (h1);
	\draw[stateTransition] (i2) -- (h2);
	\draw[stateTransition] (i2) -- (h3);
	\draw[stateTransition] (i2) -- (h4);
	\draw[stateTransition] (i2) -- (h5);
	\draw[stateTransition] (i3) -- (h1);
	\draw[stateTransition] (i3) -- (h2);
	\draw[stateTransition] (i3) -- (h3);
	\draw[stateTransition] (i3) -- (h4);
	\draw[stateTransition] (i3) -- (h5);
	
	\draw[stateTransition] (h1) -- (o1);

	\draw[stateTransition] (h2) -- (o1);

	\draw[stateTransition] (h3) -- (o1);

	\draw[stateTransition] (h4) -- (o1);

	\draw[stateTransition] (h5) -- (o1);
	\draw[stateTransition] (xy) -- (p);
	\draw[stateTransition] (p) -- (z);

	\node[above=1.3em of p, align=right] (t) {$\hat{\sigma}(\textbf{D}^{-1/2}_{\hat{\textbf{G}}}\hat{\textbf{G}}\textbf{D}^{-1/2}_{\hat{\textbf{G}}}\textbf{X}\textbf{W}+\textbf{b})$};
	
	\draw[stateTransition] (o1) -- node[above] {$\hat{\textbf{G}}$} (10.3, 0.0);
\end{tikzpicture}
}
\vspace{-3mm}
\caption{A single pathfinder neuron (\ref{fig:neuron}) and a pathfinder discovery network (\ref{fig:deep_model}) with multiple pathfinder neurons in a single hidden layer. The pathfinder graph $\tilde{\textbf{G}}$ output by the pathfinder layer is used by some graph convolutional layer $\hat{\sigma}(\cdot)$. We illustrate this here using the GCN model \cite{kipf2017semi}, showing how a learned graph can be normalized.
}
\end{figure}
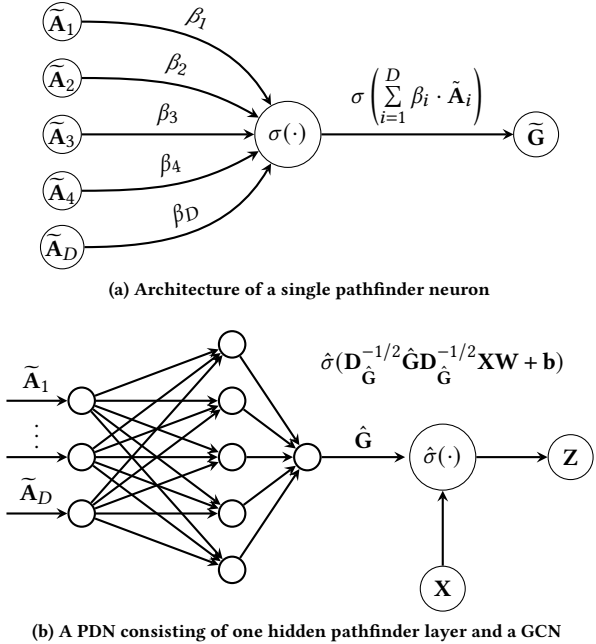

\begin{definition}
Pathfinder Discovery Networks. \emph{This general combined design of a message passing model and pathfinder layers (or neurons) is a \textit{Pathfinder Discovery Network}. An instance where the the pathfinder network has a single hidden layer is depicted in Figure \ref{fig:deep_model}}.
\end{definition}

While we focus on specific applications in this work, we note that PDNs can be used with most graph neural network models and objectives (both supervised and unsupervised).

\section{Advantages of PDNs}
As noted by \citet{grale} and others \cite{de2013influence,wu2019quest},
the performance of graph learning systems can vary greatly based on the quality of the network used.
In this section we compare methods based on the popular graph attention network (GAT) model with PDNs. While GAT is not explicitly motivated by the problem of graph construction, we note that in the current literature the GAT approach can be viewed as attempting to learn a graph jointly with a deep learning task. However, GAT and related attention models have the following critical weaknesses that make them ill-suited for graph building:
\begin{enumerate}
    \item  GAT models learn a single aggregation over a single source of features, which significantly constrains the expressivity of the graph it can learn.
    \item The GAT framework is heavily dependent on multiple attention heads for regularization. Unfortunately, the increase in parameters results in overfitting, which raises issues when training on real world datasets \cite{wang2019improving,shchur2018pitfalls}.
\end{enumerate}
\noindent
PDNs can mitigate these weaknesses.

\subsection{Expressivity}

PDNs are designed from the ground up to handle an arbitrary number of modalities, each defined as a similarity measure over the vertices. The pathfinder network is able to combine these similarity measures in arbitrary ways. 

\subsubsection{Exclusive Or}
As a motivating example, consider an XOR operation. In a multiplex graph setting, an XOR can describe a case where the presence of two edges together has a different semantic meaning than the presence of either edge separately.
\begin{table}[h!]
\caption{The PDN model is able to learn complex relationships over multiplex edges. Here we show how an XOR relationship can be learned over two networks ($\textbf{A}'_{u,v}$ and  $\textbf{A}''_{u,v}$).
\vspace{-5mm}	
}\label{tab:edge_types}
\begin{tabular}{cc|ccc}
\multicolumn{2}{c}{Edge State} & \multicolumn{3}{c}{PDN activations} \\
\hline
$\textbf{A}'_{u,v}$ & $\textbf{A}''_{u,v}$ & $h_1$&$h_2$&$\alpha_{u,v}$ \\ \hline
0    & 0    &   0&0& 0        \\
0    & 1    &   1&0& 1       \\
1    & 0    &   1&0& 1       \\
1    & 1    &   2&1& 0       \\ \hline
\end{tabular}
	
\end{table}

\noindent
\emph{Proposition}. PDNs can learn XOR operations over different layers of a multiplex graph.

\noindent
\emph{Proof}. Let us consider a node classification problem with two component binary valued edge weight vectors $(\textbf{A}'_{u,v} ,\textbf{A}''_{u,v})$ on each edge. 
Further, consider a PDN composed of a hidden layer with 2 neurons, described by these equations:
\begin{align*}
    h_1&=\text{ReLU}(\textbf{A}'_{u,v}+\textbf{A}''_{u,v})\\
    h_2&=\text{ReLU}(\textbf{A}'_{u,v}+\textbf{A}''_{u,v}-1)\\
    \alpha_{u,v}&=h_1-2\cdot h_2
\end{align*}
\noindent
As shown in Table \ref{tab:edge_features}, this network reproduces the \emph{exclusive-or} function.  We also include the hidden states ($h_1,h_2$) and predicted edge weights ($\alpha_{u,v}$) obtained with this pathfinder layer. \hfill$\blacksquare$

In practice without feature engineering GCN and GAT models can only utilize one of the edge features or the edge existence as an edge weight. Such weights on their own cannot separate different edge types.

\subsubsection{Different Edge Weight Semantics}
Similar analysis can be done when considering the case where a node has two edge types where one edge denotes similarity and one denotes distance. A PDN can correctly learn to invert the distance edge, producing a single similarity measure that incorporates the full range of provided information. Because GAT uses a softmax aggregation, it cannot learn inversions; the best it can do is ignore the distance edge. The standard GCN treats all edges as either similarity or distance, and so will end up misinterpreting the information being provided by one of the two edge types. 

\subsection{Resilience to skewed degree distributions}

Many GNN implementations suffer when faced with nodes that have very large degrees. 
As shown by \cite{anonymous2021learning} the limiting behaviour of $\alpha_{u,v}$ in the graph attention (GAT) model \cite{gat_iclr18} forces edges weights to 0 as the neighbourhood size of $u$ increases:
 \begin{align}\lim_{|N(u)|\to \infty}\alpha_{u,v}=\lim_{|N(u)|\to\infty}\frac{\exp(f_\theta(\textbf{H}_{u,:};\textbf{H}_{v,:}))}{\sum\limits_{w\in N(u)}\exp(f_\theta(\textbf{H}_{u,:};\textbf{H}_{w,:}))}=0.\label{gatlmit}
 \end{align}
This over-smoothing limits the effectiveness of GAT and similar models on real world graphs, where node degrees often follow a power law distribution \cite{anonymous2021learning}. By comparison, PDNs can score edge weights independently, and therefore do not incorrectly penalize high degree nodes.

\noindent
\emph{Proposition}. PDNs can be constructed such that high degree nodes do not drive edge weights to 0.

\noindent
\emph{Proof}. Let us consider an edge $(u,v)$ of the undirected graph $\mathcal{G}$ used for message passing. Let us denote the message passing weight of this edge as $\alpha_{u,v}$. Moreover, let us assume that the edge has a two dimensional feature vector $(\textbf{A}'_{u,v} ,\textbf{A}''_{u,v})$. 
Let us further define the behaviour of $\alpha_{u,v}$ in a PDN which has no hidden layer and has a linear activation function:
\begin{align}
\lim_{|N(u)|\to \infty}\alpha_{u,v}=\beta_1 \cdot \textbf{A}'_{u,v}+\beta_2 \cdot \textbf{A}''_{u,v}.\label{pdnlmit}
\end{align}
\noindent
In Equation \eqref{pdnlmit} the $\beta$ values are trainable parameters of the PDN. We see that the limiting behaviour of the PDN edge weight does not depend on the neighbourhood size. \hfill$\blacksquare$

High degree nodes in the GAT and GCN model will have a large number of edge weights close to zero. This results in poor quality neighbourhood representations which are not discriminative on the downstream task.

\subsection{Edge-weight  calculation time complexity}

GAT implementations are heavily regularized through averaging of multiple attention heads; experimental results in \cite{gat_iclr18} use 8 unique attention heads over all of the node features. This can result in significant overfitting for cases where there are few features, or extremely large matrix calculations for large feature spaces.

This has an impact on the runtime necessary to calculate each edge, both during training and during inference. Naive implementations of the GCN and GAT models which do not use a cache can calculate the weight $\alpha_{u,v}$ in $\mathcal{O}(|\mathcal{N}(v)|)$ time. By contrast the same edge weight can be calculated in $\mathcal{O}(1)$ time with a shallow PDN which has a linear activation function.




\section{Variations on the basic model}\label{sec:variations}
In order to understand the core motivations behind PDNs, we have thus far limited our discussion to high level, general characteristics. In this section, we drill down to specific variations on the basic model to demonstrate the flexibility and expressive power of the proposed framework.

\subsection{A model with learned similarities}
\label{subsec:edgeconv}

We have mentioned in passing that one can design a PDN where the weights in the adjacency matrices describing the similarity graphs are themselves parametrized by neural networks.  We have thus far assumed that the weights described by the adjacency matrices $\widetilde{\textbf{A}}_1,\dots, \widetilde{\textbf{A}}_D$ are coming from pre-calculated similarities. Instead, let us assume that for each \textit{binary} $\widetilde{\textbf{A}}_1,\dots, \widetilde{\textbf{A}}_D$ we have a feature matrix $\textbf{X}_i, \dots, \textbf{X}_D$.
We can then define a graph convolutional model where the edge weights of an input graph are learned by node features. Let $\textbf{H}_i$ be the node hidden representation matrix,
\begin{align}
\textbf{H}_i &= \sigma(\textbf{X}_i \cdot \textbf{W}'_{i}+\textbf{b}_{i}).\label{eq:edge_rep}
\end{align}

\noindent Here $\textbf{X}_i$ is the $i^{th}$ generic node feature matrix, the function $\sigma(\cdot)$ is an elementwise non-linearity, and $\textbf{W}'_{i}$ and $\textbf{b}_{i}$ are the feature matrix specific trainable weight matrix and bias vector. Using the endpoint representations we define $\hat{\textbf{G}}_{i}$ as a learned input adjacency matrix for a graph learning neuron:
\begin{align}
\hat{\textbf{G}}_{i}&=\textbf{A}_i \odot \hat{\sigma}(\textbf{H}_i\cdot \textbf{H}_{i}^{\top}).\label{eq:edge_score}
\end{align}
We use the elementwise non linearity $\hat{\sigma}(\cdot)$ to transform the raw edge weights which are conditioned on the original adjacency matrix by a Hadamard product. Exploiting the similarity of the individual adjacency matrices the calculation of \eqref{eq:edge_score} happens in $\mathcal{O}(|E|)$.

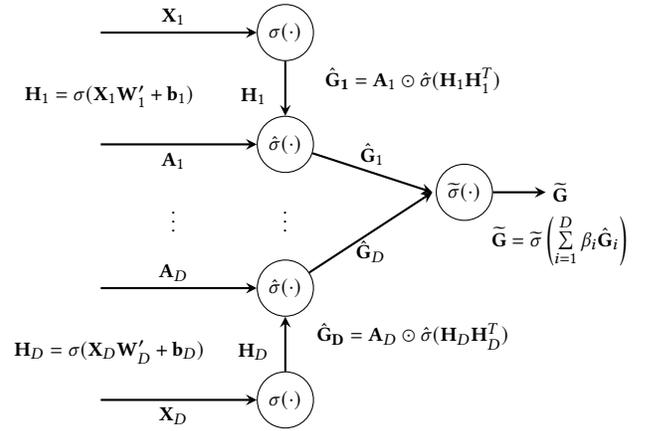
\begin{figure}[h!]

\begin{tikzpicture}[scale=0.85,transform shape]

	\node[draw,circle,minimum size=25pt,inner sep=0pt] (i1) at (8, 1.0) {$\sigma (\cdot )$};
	\node[draw,circle,minimum size=25pt,inner sep=0pt] (o1) at (8,-0.75) {$\hat{\sigma}(\cdot)$};
	
	\node[draw,circle,minimum size=25pt,inner sep=0pt] (ap) at (10.8,-1.5) {$\widetilde{\sigma}(\cdot)$};
	\node[] (w) at (6.25,1.25) {$\textbf{X}_1$};	
	\node[] (w) at (6.25,-1.0) {$\textbf{A}_1$};	
	\node (inp1) at (5, 1.0) {};
	\node (inp2) at (5,-0.75) {};

	\draw[stateTransition] (inp1) -- (i1);
	\draw[stateTransition] (inp2) -- (o1);
	
	\draw[stateTransition] (i1) -- (o1);
	
	\node[] (pp) at (5.25,0.0)  {$\textbf{H}_1 = \sigma (\textbf{X}_1\textbf{W}'_1+\textbf{b}_{1})$};	
	\node[] (w) at (7.5,-0.0) {$\textbf{H}_1$};

	\node[] (tk) at (10,0.25) {$\hat{\textbf{G}}_{\textbf{1}}=\textbf{A}_1\odot \hat{\sigma}(\textbf{H}_1 \textbf{H}_1^T)$};	
	
	\draw[stateTransition] (o1) -- node[above] {$\hat{\textbf{G}}_1$} (10.3, -1.5);

	\node[draw,circle,minimum size=25pt,inner sep=0pt] (ip1) at (8, -3.0) {$\hat{\sigma} (\cdot )$};
	\node[draw,circle,minimum size=25pt,inner sep=0pt] (op1) at (8,-4.75) {$\sigma(\cdot)$};
	\node[] (wp) at (6.25,-5.0) {$\textbf{X}_D$};	
	\node[] (wp) at (6.25,-2.75) {$\textbf{A}_D$};	
	\node (inpp1) at (5, -3.0) {};
	\node (inpp2) at (5,-4.75) {};

	\draw[stateTransition] (inpp1) -- (ip1);

	\draw[stateTransition] (inpp2) -- (op1);
	
	\draw[stateTransition] (op1) -- (ip1);
	
	\node[] (ppp) at (5.25,-4.0)  {$\textbf{H}_D = \sigma (\textbf{X}_D\textbf{W}'_D+\textbf{b}_{D})$};	
	\node[] (wp) at (7.5,-4.0) {$\textbf{H}_D$};
		
	\draw[stateTransition] (ip1) -- node[below] {$\hat{\textbf{G}}_D$} (10.3, -1.5);
	\node[] (agg) at (12.3,-1.5) {$\widetilde{\textbf{G}}$};
	\node[] (wp) at (6.25,-1.85) {$\vdots$};
	\node[] (wp) at (8.0,-1.85) {$\vdots$};
	
	\node[] (tk) at (10,-3.75) {$\hat{\textbf{G}}_{\textbf{D}}=\textbf{A}_D\odot \hat{\sigma}(\textbf{H}_D \textbf{H}_D^T)$};
	
	\node[] (tk) at (12.3,-2.25) {$\widetilde{\textbf{G}}=\widetilde{\sigma}\left(\sum\limits_{i=1}^D \beta_i \hat{\textbf{G}}_i\right)$};
	
	\draw[stateTransition] (ap) -- (agg);	
\end{tikzpicture}
\caption{The pathfinder neuron design with learned similarity scores. From each node feature matrix conditioned by the corresponding adjacency matrix we learn a similarity graph. In the pathfinder neuron we learn to combine these together as a single learned graph denoted by $\tilde{\textbf{G}}$. This output graph can serves as the input for an arbitrary downstream graph convolutional layer.}\label{fig:edge_conv}
\end{figure}

A pathfinder neuron receives multiple learned graphs as input, combines those and outputs a final graph. This idea is summarized by Figure \ref{fig:edge_conv} where we have $D$ different feature matrices and from each of them we learn a separate graph that we use as input for the pathfinder neuron, which in turn outputs $\widetilde{\textbf{G}}$. This final aggregation is defined by Equation \eqref{eq:grale_edge_conv_neuron} in which $\beta_i$ is a learned parameter that acts as a weight for the learned graphs and $\widetilde{\sigma}(\cdot)$ is a non-linearity. 
\begin{align}
\widetilde{\textbf{G}}&=\widetilde{\sigma}\left(\sum\limits_{i=1}^D\beta_i\hat{\textbf{G}}_i \right)\label{eq:grale_edge_conv_neuron}
\end{align}

\subsection{A model with cheap multi-scale mixing}
\label{subsec:multiscale}
Multi-scale graph neural network models obtain information about the neighbourhoods of nodes at multiple hops \cite{perozzi2017don, mixhop_icml19, rozemberczki2019multiscale, feather} and learn features for each hop. Most graph neural networks \cite{kipf2017semi,graphsage_nips17, gwnn_iclr19, clustergcn_kdd19, ppnp_iclr19} which are not multi-scale (with the exception being \textit{AttentionWalk} \cite{wus_nips18} and \textit{DCRNN} \cite{dcrnn}) pool features from neighbourhoods at different scales without considering what is the optimal mixing of information. In the following we will define a corner case of our model which allows for supervised and explainable pooling of multi-scale information with trainable weights.
\begin{center}
\begin{algorithm}[h!]
\DontPrintSemicolon
\SetAlgoLined
    \KwData{$\widetilde{\textbf{A}}$ - Normalized adjacency matrix\\
\quad \quad\,\,\,\,\,$\textbf{X}$ - Feature matrix\\
\quad \quad\,\,\,\,\,$D$ - Order of adjacency matrix powers\\
\quad \quad\,\,\,\,\,$d$ - Number of filters}
    \KwResult{$\textbf{Z}$ -- Hidden state matrix}
    $\textbf{Z}\leftarrow$ Initialize representations(\textit{d}).\;
    $\textbf{Z}_{0}\leftarrow$ \textbf{X}\textbf{W}\;    
    \For{$i \in \left\{1,\dots,D\right \}$}{

        $\textbf{Z}_i \leftarrow \widetilde{\textbf{A}}\textbf{Z}_{i-1} $\;

    $\textbf{Z}\leftarrow \textbf{Z}+P_i\cdot \textbf{Z}_i$\;
    }
\caption{\textbf{Efficient sparsity aware forward pass multi-scale mixing with a softmax learned graph and a linear graph convolutional activation function.}}\label{alg:multi_scale_forward_pass}
\end{algorithm}
\end{center}

Let $\widetilde{\textbf{A}}$ be the normalized adjacency matrix of the weighted undirected graph $\textbf{G}$. We assume that the similarity graphs of interest are described by powers of this normalized adjacency matrix for a given $D$ number of hops -- 
$\widetilde{\textbf{A}}_i=\widetilde{\textbf{A}}^i,\quad \forall i=1,\dots,D$. The learned graph used for the forward pass is defined as:
\begin{align}
\widehat{\textbf{G}} &=\sum \limits_{i=1}^D P_i\cdot \widetilde{\textbf{A}_i}\label{eq:softmax_multi}
\end{align}
where $P_i$ is the weight of a given adjacency matrix power, and is parametrized with a softmax as $\exp(\alpha_i)/\left (\sum_{i=1}^{D} \exp(\alpha_i)\right)$. As the direct calculation of the adjacency matrix powers is prohibitive, we instead use an efficient forward pass algorithm to calculate the hidden state matrices described by Algorithm \ref{alg:multi_scale_forward_pass}. The core idea is to exploit the sparsity of the adjacency matrix in each iteration by using the normalized adjacency to average node representations, and weighting the representations with learned $P_i$ scores.

\section{Experiments}\label{sec:experiments}
Above, we theoretically motivated the development of PDNs by discussing the importance of jointly learning graphs and GNNs for specific tasks and evaluating the expressivity of the pathfinder layers. In the following, we empirically validate our analysis by demonstrating that PDNs have a significant advantage on a class of graph learning tasks, while maintaining competitive predictive performance on other baselines. We also describe how weights in a pathfinder neuron can be interpreted as attention, and we analyze model runtime to discuss the scalability of our models.

\subsection{Synthetic node classification experiment}\label{subsec:synth}
PDNs are a natural fit for dealing with noisy node and edge features, because they can learn complex correlations across many different combined modalities of data while removing unimportant information. To highlight this key advantage, we investigate node classification performance on synthetically generated datasets that are specifically designed with imperfect feature information. The detailed settings of the synthetic node classification experiments are discussed in Appendix \ref{app:synthetic_settings}.
\begin{table}[h!]
\caption{Synthetic node classification scenarios with the range of the manipulated hyperparameters and specific implications of the modulation in the scenario.}\label{tab:scenarios}
\begin{tabular}{c ll}
\hline
\textbf{Scenario} & \textbf{Parameter} &  \textbf{Implication of increase} \\ \hline
1        &     $C\in[2,6]$    &           Less clear classes             \\
2        & $n\in[2^4,2^{10}]$          & More instances for generalization\\
3        &     $P\in[2^{-10},2^{-4}]$                  &  Stronger class cohesion           \\
4        & $Q\in[2^{-10},2^{-4}] $                       &     More inter-class edges      \\
5        &   $F\in[2^2,2^6]$                   & More node features            \\
6        &  $D\in[2^2,2^6]$              &    More edge features     \\
7        &  $\sigma_F\in[2^{-1},2^5]$  & Lower node feature quality\\
8        &   $\sigma_D\in[2^{-1},2^5]$  & Easier separation of edge type     \\ \hline
\end{tabular}
\end{table}

\noindent
\textbf{Synthetic data generation algorithm.}
Each synthetic graph has $C$ node label classes and $n$ nodes in the graph belonging to a given class. These two hyperparameters decide the overall number of nodes in the synthetically generated graph, $C \times n$. We generate features as follows:
\begin{enumerate}
    \item \textit{Generation of correlated node features.} For each node we generate $d_N$ continuous node features which are standard normally distributed with a pre-defined correlation structure. The eigenvalues of the node feature correlation matrix are distributed proportional to a standard half-normal distribution. This ensures that the eigenvalues of the generated correlation matrix are positive. 
    \item \textit{Generation of node labels.} The node feature matrix $\textbf{X}$ is multiplied by a normally distributed $F$ dimensional weight vector $\textbf{w}$ which results in a continuous node target feature \textbf{y}. We add zero mean normally distributed noise to this target vector with standard deviation $\sigma_F$ which results in the noisy target vector $\widetilde{\textbf{y}}$. We quantile bin the continuous target vector to get a label vector for the node classification task with $C$ distinct classes.     
    \item \textit{Edge addition.} We define two edge types in our graph: intra-class edges (those edges between nodes that share a class); and inter-class edges (the opposite). An edge exists between two intra-class nodes with probability $P$, while an edge exists between two inter-class nodes with probability $Q$.
    \item \textit{Generation of edge features.} For each edge we generate $D$ continuous edge features which are normally distributed and uncorrelated. Inter-class edge features have a standard deviation of $\sigma_D$ while intra-class edge features are distributed according to the standard normal distribution. This allows us to tune how much information can be propagated from the edges themselves.
\end{enumerate}

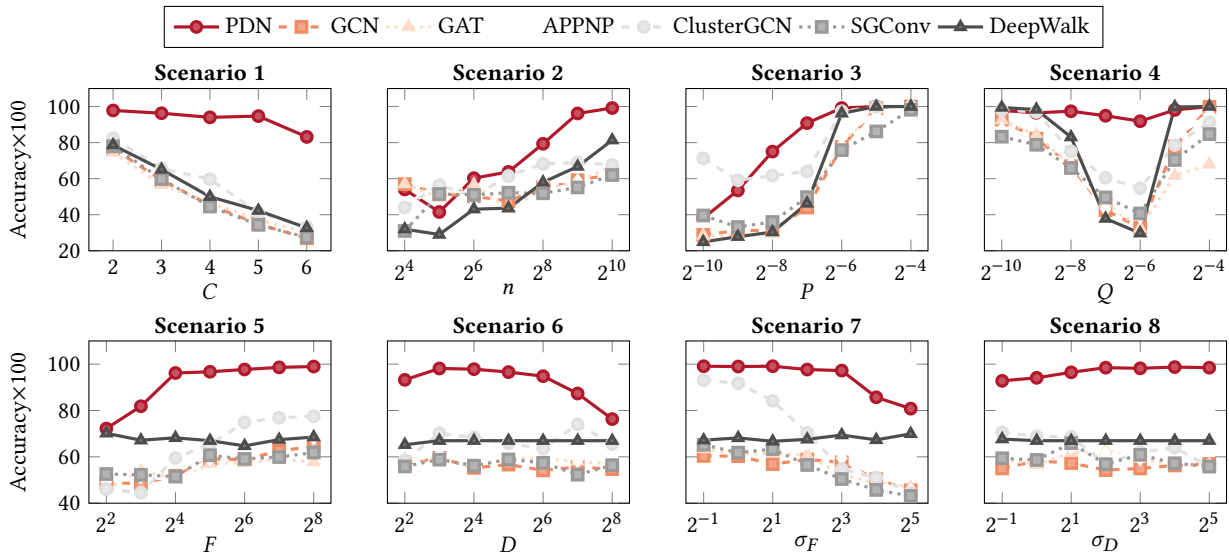
\begin{figure*}[h!]
\centering
\begin{tikzpicture}
\begin{groupplot}[group style={
                      group name=myplot,
                      group size= 4 by 2, horizontal sep=0.75cm,vertical sep=1.2cm},height=3.74cm,width=4.8cm, ymin=-10,ymax=110,ytick={0,20,40,60,80,100},title style={at={(0.5,0.9)},anchor=south},every axis x label/.style={at={(axis description cs:0.5,-0.15)},anchor=north},
    cycle list/RdGy-7,
    mark list fill={.!75!white},
    cycle multiindex* list={
                RdGy-7
                    \nextlist
                my marks
                    \nextlist
                [3 of]linestyles
                    \nextlist
                very thick
                    \nextlist
            },]
\nextgroupplot[
 	title = \textbf{Scenario 1},
 	legend columns=7,
	legend style={at={(2.25,1.25)},anchor=south},
    legend entries={PDN, GCN, GAT, APPNP, ClusterGCN, SGConv, DeepWalk},
	ylabel=Accuracy$\times100$,
	xlabel=$C$,
	xtick={2,3,4,5,6},
	xmin=1.5,
	xmax=6.5,
	ymin=20,
]
\addplot coordinates {
(2,97.85)
(3,96.3)
(4,94.02)
(5,94.72)
(6,83.17)

};
\addplot coordinates {
(2,76.75)
(3,59.17)
(4,45.32)
(5,34.78)
(6,26.95)

};
\addplot coordinates {
(2,74.55)
(3,56.67)
(4,46.7)
(5,37.28)
(6,30.08)
};
\addplot coordinates {(2,79.9)
(3,64.5)
(4,53.0)
(5,40.4)
(6,31.8)};
\addplot coordinates {
(2,82.45)
(3,65.6)
(4,59.75)
(5,41.22)
(6,33.23)
};
\addplot coordinates {
(2,77.95)
(3,59.77)
(4,44.52)
(5,34.38)
(6,27.47)
};
\addplot coordinates {
(2,78.64)
(3,65.09)
(4,50.0)
(5,42.3)
(6,32.7)};
\nextgroupplot[
 	title = \textbf{Scenario 2},
	yticklabels={,,},
	xlabel=$n$,
	xtick={16,64,256,1024},
	xmode=log,
	xmin=11.31,
	xmax=1448,
	ymin=20,	
	log basis x={2}
	]
\addplot coordinates {
(16,54.0)
(32,41.5)
(64,60.26)
(128,63.77)
(256,79.35)
(512,96.17)
(1024,99.2)

};

\addplot coordinates {
(16,57.0)
(32,51.5)
(64,50.26)
(128,47.53)
(256,55.26)
(512,59.09)
(1024,62.36)

};

\addplot coordinates {
(16,57.0)
(32,49.5)
(64,56.67)
(128,43.12)
(256,53.9)
(512,58.44)
(1024,66.88)

};

\addplot coordinates {
(16,45.0)
(32,58.0)
(64,48.21)
(128,57.92)
(256,58.18)
(512,58.6)
(1024,72.34)

};


\addplot coordinates {
(16,44.0)
(32,56.5)
(64,50.26)
(128,61.43)
(256,68.18)
(512,68.99)
(1024,67.71)

};

\addplot coordinates {
(16,31.0)
(32,51.5)
(64,51.03)
(128,52.21)
(256,51.88)
(512,55.16)
(1024,62.03)

};

\addplot coordinates {
(16,32.0)
(32,29.0)
(64,43.08)
(128,43.64)
(256,57.99)
(512,66.72)
(1024,81.37)

};

\nextgroupplot[
 	title = \textbf{Scenario 3},
	yticklabels={,,},
	xlabel=$P$,
	xmode=log,
	xtick={0.0009765625 ,0.00390625, 0.015625,0.0625},	
	log basis x={2},
	xmin=0.000690,
	xmax=0.088,
	ymin=20,	
]
 \addplot coordinates {
 (0.0009765625,38.07)
(0.001953125,53.47)
(0.00390625,75.03)
(0.0078125,90.83)
(0.015625,99.2)
(0.03125,100.0)
(0.0625,100.0)
};
 \addplot coordinates {
 (0.0009765625,28.83)
(0.001953125,31.2)
(0.00390625,31.33)
(0.0078125,44.00)
(0.015625,77.67)
(0.03125,99.1)
(0.0625,100.0)
};

 \addplot coordinates {
(0.0009765625,28.43)
(0.001953125,28.57)
(0.00390625,29.53)
(0.0078125,47.03)
(0.015625,79.7)
(0.03125,98.9)
(0.0625,100.0)

};

 \addplot coordinates {
(0.0009765625,36.1)
(0.001953125,32.73)
(0.00390625,34.83)
(0.0078125,53.23)
(0.015625,83.3)
(0.03125,96.67)
(0.0625,99.77)

};

 \addplot coordinates {
(0.0009765625,71.27)
(0.001953125,59.13)
(0.00390625,61.7)
(0.0078125,63.97)
(0.015625,97.43)
(0.03125,99.83)
(0.0625,100.0)

};

 \addplot coordinates {

(0.0009765625,39.57)
(0.001953125,33.17)
(0.00390625,35.97)
(0.0078125,49.77)
(0.015625,75.83)
(0.03125,86.23)
(0.0625,98.23)

};

 \addplot coordinates {
(0.0009765625,24.93)
(0.001953125,27.8)
(0.00390625,30.37)
(0.0078125,46.27)
(0.015625,96.23)
(0.03125,100.0)
(0.0625,100.0)

};

\nextgroupplot[
 	title = \textbf{Scenario 4},
	yticklabels={,,},
	xlabel=$Q$,
	xmode=log,
	xtick={0.0009765625 ,0.00390625, 0.015625,0.0625},	
	log basis x={2},
	xmin=0.000690,
	xmax=0.088,
	ymin=20,	
]
 \addplot coordinates {
(0.0009765625,97.83)
(0.001953125,96.53)
(0.00390625,97.4)
(0.0078125,94.9)
(0.015625,91.87)
(0.03125,98.13)
(0.0625,100.0)

};
 \addplot coordinates {
(0.0009765625,92.77)
(0.001953125,82.23)
(0.00390625,66.9)
(0.0078125,42.57)
(0.015625,32.73)
(0.03125,77.97)
(0.0625,99.87)

};

 \addplot coordinates {
(0.0009765625,92.43)
(0.001953125,84.3)
(0.00390625,67.43)
(0.0078125,46.4)
(0.015625,32.4)
(0.03125,61.7)
(0.0625,67.93)

};

 \addplot coordinates {
(0.0009765625,87.2)
(0.001953125,76.93)
(0.00390625,68.2)
(0.0078125,43.23)
(0.015625,38.87)
(0.03125,83.17)
(0.0625,95.87)

};

 \addplot coordinates {
(0.0009765625,98.2)
(0.001953125,96.97)
(0.00390625,75.17)
(0.0078125,60.5)
(0.015625,54.73)
(0.03125,78.67)
(0.0625,91.27)

};

 \addplot coordinates {
(0.0009765625,83.3)
(0.001953125,78.83)
(0.00390625,65.93)
(0.0078125,49.57)
(0.015625,40.7)
(0.03125,70.37)
(0.0625,84.7)
};

 \addplot coordinates {
(0.0009765625,99.47)
(0.001953125,98.37)
(0.00390625,83.07)
(0.0078125,37.87)
(0.015625,29.73)
(0.03125,99.77)
(0.0625,100.0)
};

\nextgroupplot[
 	title = \textbf{Scenario 5},
 		ylabel=Accuracy$\times100$,
	xlabel=$F$,
	xmode=log,
	xmin=2.82,
	xmax=362,
	xtick={4,16,64,256},
	ymin=40,	
	log basis x={2}
]
 \addplot coordinates {
 (4,72.2)
(8,81.83)
(16,96.17)
(32,96.7)
(64,97.7)
(128,98.6)
(256,98.97)
};
  \addplot coordinates {
(4,48.77)
(8,48.53)
(16,51.8)
(32,59.13)
(64,58.83)
(128,62.57)
(256,64.53)  
};

  \addplot coordinates {
(4,46.0)
(8,53.8)
(16,50.77)
(32,57.1)
(64,57.47)
(128,59.07)
(256,57.8)
};

  \addplot coordinates {
(4,49.13)
(8,51.47)
(16,50.07)
(32,59.23)
(64,68.77)
(128,70.47)
(256,68.2)

};

  \addplot coordinates {
(4,46.07)
(8,44.5)
(16,59.53)
(32,65.43)
(64,74.87)
(128,76.9)
(256,77.47)

};

  \addplot coordinates {
(4,52.6)
(8,52.23)
(16,51.47)
(32,60.7)
(64,59.1)
(128,59.93)
(256,61.83)
};

  \addplot coordinates {
(4,70.17)
(8,67.17)
(16,68.2)
(32,66.97)
(64,64.73)
(128,67.43)
(256,68.53)
};

\nextgroupplot[
 	title = \textbf{Scenario 6},
	yticklabels={,,},
	xlabel=$D$,
	xmode=log,
	xmin=2.82,
	xmax=362,
	xtick={4,16,64,256},
	ymin=40,		
	log basis x={2}
]
 \addplot coordinates {
 (4,93.23)
(8,98.13)
(16,97.83)
(32,96.57)
(64,94.8)
(128,87.33)
(256,76.3)
};
  \addplot coordinates {
 (4,56.93)
(8,59.57)
(16,55.27)
(32,56.67)
(64,54.17)
(128,55.63)
(256,54.7)

};

\addplot coordinates {
(4,58.93)
(8,59.33)
(16,56.1)
(32,59.3)
(64,59.27)
(128,58.07)
(256,56.77)

};

\addplot coordinates {
(4,57.83)
(8,61.3)
(16,62.73)
(32,61.6)
(64,62.1)
(128,58.5)
(256,62.47)
};

\addplot coordinates {
(4,58.47)
(8,70.13)
(16,68.4)
(32,65.77)
(64,63.57)
(128,74.03)
(256,65.57)
};

\addplot coordinates {
(4,55.87)
(8,58.8)
(16,56.17)
(32,58.87)
(64,57.43)
(128,52.3)
(256,56.4)

};

\addplot coordinates {
(4,65.23)
(8,66.97)
(16,66.97)
(32,66.97)
(64,66.97)
(128,66.97)
(256,66.97)

};
\nextgroupplot[
 	title = \textbf{Scenario 7},
	yticklabels={,,},
	xlabel=$\sigma_F$,
	xmode=log,
	xmin=0.35,
	xmax=45.25,
	xtick={0.5,2,8,32},
	ymin=40,		
	log basis x={2}
]
 \addplot coordinates {
(0.5,99.1)
(1,98.97)
(2,99.1)
(4,97.67)
(8,97.23)
(16,85.7)
(32,80.8)

 };
  \addplot coordinates {
(0.5,60.4)
(1,60.3)
(2,56.87)
(4,58.87)
(8,57.63)
(16,50.4)
(32,45.6)

  };
  
\addplot coordinates {

(0.5,63.43)
(1,61.17)
(2,61.47)
(4,60.83)
(8,52.2)
(16,48.83)
(32,46.87)

};
\addplot coordinates {

(0.5,74.8)
(1,72.97)
(2,74.47)
(4,66.17)
(8,59.5)
(16,48.43)
(32,45.33)

};
\addplot coordinates {

(0.5,93.1)
(1,91.73)
(2,84.17)
(4,70.37)
(8,54.7)
(16,51.03)
(32,44.57)

};
\addplot coordinates {

(0.5,65.2)
(1,61.83)
(2,63.37)
(4,56.5)
(8,50.43)
(16,45.73)
(32,43.13)

};

\addplot coordinates {

(0.5,67.2)
(1,68.17)
(2,66.7)
(4,67.57)
(8,69.4)
(16,67.33)
(32,69.93)

};

\nextgroupplot[
 	title = \textbf{Scenario 8},
	yticklabels={,,},
	xlabel=$\sigma_D$,
	xmode=log,
	xmin=0.35,
	xmax=45.25,
	xtick={0.5,2,8,32},
	ymin=40,		
	log basis x={2}
]
 \addplot coordinates {
 (0.5,92.8)
(1,94.1)
(2,96.47)
(4,98.47)
(8,98.2)
(16,98.73)
(32,98.47)
 };
  \addplot coordinates {
 (0.5,55.0)
(1,58.27)
(2,57.33)
(4,54.3)
(8,54.97)
(16,56.3)
(32,56.93)

  };
  \addplot coordinates {
(0.5,58.97)
(1,56.67)
(2,61.73)
(4,62.67)
(8,59.2)
(16,56.27)
(32,56.93)

  };
   \addplot coordinates {
(0.5,65.27)
(1,61.87)
(2,64.3)
(4,61.77)
(8,59.23)
(16,61.4)
(32,65.23)

  };  
    \addplot coordinates {
(0.5,70.6)
(1,68.8)
(2,68.6)
(4,57.07)
(8,62.07)
(16,64.0)
(32,55.73)

  };  
   \addplot coordinates {
(0.5,59.37)
(1,58.67)
(2,65.9)
(4,56.83)
(8,60.87)
(16,57.23)
(32,55.8)

  }; 
    \addplot coordinates {
(0.5,67.67)
(1,66.97)
(2,66.97)
(4,66.97)
(8,66.97)
(16,66.97)
(32,66.97)
  };  
  
\end{groupplot}
\end{tikzpicture}

\caption{Node classification performance measured by average test set accuracy (10 experimental repetitions) on the synthetically generated attributed graphs for the scenarios described in Table \ref{tab:scenarios}. The proposed Pathfinder Discovery Network architecture has robust predictive performance under a wide range of synthetic data generation hyperparameters.}\label{fig:scenarios}

\end{figure*}

 \textbf{Findings and Discussion.}
The mean accuracy scores for each scenario are shown in Figure \ref{fig:scenarios}. PDN materially outperforms the baselines for a wide range of synthetic graphs. The results of Scenario 1 demonstrate that PDN is able to distinguish between less clearly defined classes, while competing graph neural networks struggle to maintain competitive performance.  We highlight data efficiency in Scenario 2 -- given a fixed number of instances, PDN generalizes better to unseen data where $n\geq 2^6$. In Scenario 3, we observe that stronger class-cohesion results in better classification performance for all models, and that PDN displays superior marginal predictive performance gains. As one can see ClusterGCN uses a pre-processing step which is purely topological and this filters out inter-class edges. Increasing the number of inter-class edges in Scenario 4 initially decreases the predictive performance of the baselines; by comparison, PDN is able to learn to ignore the noise propagating inter-class edges. Scenario 5 shows that all supervised models gain when more vertex features are available, and again PDN displays superior marginal performance gain. On the contrary, we see that the PDN overfits when a large number of edge features is available based on Scenario 6. Though all models are sensitive to node features, Scenario 7 shows that PDNs are significantly more resilient to node feature corruption. Finally, in Scenario 8, higher quality edge features only help PDN.

We briefly want to focus on the results in Scenario 4, as we believe this demonstrates the XOR functionality of the PDN. The GCN baseline models learn from the expected value of neighboring hidden states. In high homophily graphs (i.e. the low Q region), neighboring states will correlate with node features resulting in high performance. The same holds for low homophily graphs (i.e. the high Q region), except the weights are inverted -- in other words, the baseline GCN models will learn to simply invert neighboring node states. In the middle, the baseline models cannot learn a single aggregation that correctly handles the differing edge information. By contrast, PDNs are expressive enough to learn to differentiate the edge weights, allowing it to maintain high performance throughout.
\input{figures/tsne_synth}

 \textbf{Implicit learning of inter- and intra-class edges.} We perform a visual embedding analysis in Figure \ref{fig:tsne}, where we examine the 2 dimensional t-SNE embeddings \cite{tsne_1, tsne_2} of hidden layer edge representations for the PDN and GAT models. The representations extracted from the PDN show distinct separation for the inter and intra-class edges, which implies that the model has learned to meaningfully distinguish these two modalities of information. By contrast, the GAT representations are not separated by the type of the edge. This further demonstrates the high expressiveness of PDNs.

\subsection{Multiplex node classification performance}
We evaluated the predictive performance of PDNs on real world node classification problems using publicly available multiplex webgraph datasets \cite{dmgi}. The descriptive statistics of these graph datasets are presented Appendix \ref{app:descriptives_multiplex}.

\textbf{Experimental settings.} Our experiments focused on 100-shot node classification and we calculated the average test accuracy of multiplex graph neural network architectures. We included a range of supervised models \cite{dmgi,khan2019multi} and unsupervised proximity preserving and attributed node embedding techniques \cite{shi2018mvn2vec,matsuno2018mell,mne}. The exact experimental settings are described in Appendix  \ref{app:multiplex_settings}.
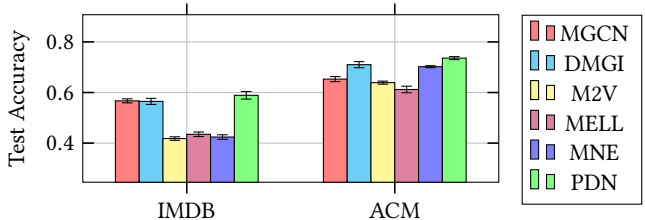
\begin{figure}[h!]
\centering
\begin{tikzpicture}
\begin{axis}[every tick/.style={black,semithick,},
width=2.8in,
height=1.5in,
xmajorgrids=true,
ymajorgrids=true,
    ybar=0pt,
    bar width=9pt,
    enlargelimits=0.5,
    legend style={at={(1.2,1.0)},
      anchor=north,legend columns=1},
    ylabel={Test Accuracy},
    symbolic x coords={IMDB, ACM},
    xtick=data,
    ]


\addplot[fill=red!50,error bars/.cd,
y dir=both,y explicit,
]  coordinates {(IMDB,0.567)+-(0.008,0.008) (ACM,0.653)+-(0.010,0.010)};

\addplot[fill=cyan!50,error bars/.cd,
y dir=both,y explicit,
]  coordinates {(IMDB,0.565)+-(0.012,0.012) (ACM,0.710)+-(0.012,0.012)};
\addplot[fill=yellow!50,error bars/.cd,
y dir=both,y explicit,
]  coordinates {(IMDB,0.418)+-(0.007,0.007) (ACM,0.639)+-(0.006,0.006)};
\addplot[fill=purple!50,error bars/.cd,
y dir=both,y explicit,
]  coordinates {(IMDB,0.435)+-(0.009,0.009) (ACM,0.612)+-(0.013,0.013)};
\addplot[fill=blue!50,error bars/.cd,
y dir=both,y explicit,
]  coordinates {(IMDB,0.424)+-(0.009,0.009) (ACM,0.702)+-(0.004,0.004)};

\addplot[fill=green!50,error bars/.cd,
y dir=both,y explicit,
]  coordinates {(IMDB,0.589)+-(0.015,0.015) (ACM,0.736)+-(0.006,0.006)};

\legend{MGCN\quad,DMGI\quad,M2V\quad,MELL\quad,MNE\quad,PDN\quad}

\end{axis}
\end{tikzpicture}
\caption{Average multiplex 100-shot node classification test accuracy results calculated from 10 experimental runs (the error bars standard deviations around the mean) on mutiplex graph benchmark datasets.}\label{fig:multiplexity}

\end{figure}

\textbf{Findings and discussions.} The average test accuracy scores are plotted on Figure \ref{fig:multiplexity} with standard deviations around the mean. Our results demonstrate that PDNs significantly outperform the competing supervised and unsupervised multiplex graph representation learning techniques on these datasets in terms of test accuracy. It is also evident that supervised learning methods have a considerable performance advantage over the unsupervised ones.

\subsection{Node classification performance}
\label{subsec:performance}
Excitingly, PDNs are quite capable in the multiplex graph settings. That said, we want to ensure PDNs maintain high performance in traditional single graph settings. Further, we believe that joint training of the pathfinder and classifier will lead to lift even on well established problems. We therefore evaluate the node classification performance of our proposed model variants on widely used citation graphs \cite{lu2003link, namata2012query} and social networks \cite{rozemberczki2019multiscale, feather}. The descriptive statistics of these datasets are in Table \ref{tab:desriptives} of Appendix \ref{app:descriptives}.

\textbf{Experimental settings.} 
Because PDNs are general, we included a wide variety of unsupervised and supervised baselines to best understand relative performance. We compared the predictive performance to node embeddings and various graph neural networks. The exact experimental settings can be found in Appendix \ref{app:real_world_settings}.

\textbf{Findings and Discussion.}
We report the mean accuracy estimates with standard deviations in Table \ref{tab:accuracy}. Our results demonstrate that PDNs outperform unsupervised methods by between 2.5 and 16.5 \% in terms of accuracy. Against all approaches, including supervised approaches, PDN variants are the most competitive models on the Cora, Pubmed, Facebook, and Deezer benchmarks, with a relative accuracy advantage between 0.8 and 3.5\%. PDNs fall behind only the standard GCN model on the Citeseer benchmark.

\begin{table}[h!]

\caption{Average node classification test accuracy results of 100-shot learning runs calculated from 10 experimental runs (standard deviations around the mean below the accuracy) on citation graph datasets and social networks. Bold red numbers denote the best performing model.}\label{tab:accuracy}

{\footnotesize
\begin{tabular}{lccccc}
\textbf{Model}& \textbf{Citeseer} &\textbf{Cora}& \textbf{Pubmed} &\specialcell{\textbf{Facebook}\\\textbf{Pages}}&\specialcell{\textbf{Deezer}\\\textbf{Europe}}\\
\hline
\textbf{LINE}$_2$  \citep{tang2015line}&$\underset{\pm 0.013}{ 0.470}$&$\underset{\pm 0.013}{ 0.686}$&$\underset{\pm 0.017}{ 0.675}$ &$\underset{\pm 0.010}{ 0.762}$&$\underset{\pm 0.005}{ 0.503}$ \\ [0.75em]
\textbf{DeepWalk} \citep{deepwalk}&$\underset{\pm 0.010}{ 0.523}$  &$\underset{\pm 0.011}{ 0.762}$ &$\underset{\pm 0.014}{ 0.704}$&$\underset{\pm 0.012}{ 0.531}$& $\underset{\pm 0.007}{ 0.510}$ \\[0.75em]
\textbf{Walklets} \citep{perozzi2017don}& $\underset{\pm 0.010}{ 0.513}$ &$\underset{\pm 0.010}{ 0.735}$& $\underset{\pm 0.017}{ 0.675}$&$\underset{\pm 0.011}{ 0.819}$& $\underset{\pm 0.008}{ 0.511}$\\[0.75em]
\textbf{GraRep} \citep{grarep}&$\underset{\pm 0.027}{ 0.421}$  &$\underset{\pm 0.016}{ 0.634}$&$\underset{\pm 0.018}{ 0.653}$  &$\underset{\pm 0.008}{ 0.705}$&$\underset{\pm 0.008}{ 0.507}$\\[0.75em]
\textbf{HOPE} \citep{hope}&$\underset{\pm 0.041}{ 0.397}$  &$\underset{\pm 0.026}{ 0.717}$&$\underset{\pm 0.035}{ 0.561}$  &$\underset{\pm 0.027}{ 0.593}$&$\underset{\pm 0.029}{ 0.508}$\\[0.75em]
\textbf{NetMF} \citep{netmf}&$\underset{\pm 0.030}{ 0.446}$  &$\underset{\pm 0.009}{ 0.707}$&$\underset{\pm 0.012}{ 0.710}$  &$\underset{\pm 0.015}{ 0.756}$&$\underset{\pm 0.010}{ 0.512}$\\[0.75em]\hline
\textbf{AANE} \citep{huang2017accelerated}&$\underset{\pm 0.009}{ 0.691}$  &$\underset{\pm 0.009}{ 0.760}$& $\underset{\pm 0.010}{ 0.801}$&$\underset{\pm 0.008}{ 0.652}$& $\underset{\pm 0.007}{ 0.621}$\\[0.75em]
\textbf{ASNE} \citep{liao2018attributed}& $\underset{\pm 0.015}{ 0.589}$ &$\underset{\pm 0.011}{ 0.758}$ &$\underset{\pm 0.017}{ 0.738}$&$\underset{\pm 0.010}{ 0.636}$&$\underset{\pm 0.011}{ 0.608}$ \\[0.75em]
\textbf{MUSAE} \citep{rozemberczki2019multiscale}& $\underset{\pm 0.012}{ 0.636}$ &$\underset{\pm 0.011}{ 0.758}$&{$\underset{\pm 0.004}{ 0.784}$}&$\underset{\pm 0.010}{ 0.822}$&$\underset{\pm 0.010}{ 0.563}$\\[0.75em]
\textbf{TADW} \citep{yang2015network} &$\underset{\pm 0.008}{ 0.657}$  &$\underset{\pm 0.009}{ 0.644}$ &$\underset{\pm 0.004}{ 0.765}$&$\underset{\pm 0.012}{ 0.536}$&$\underset{\pm 0.010}{ 0.558}$ \\[0.75em]
\textbf{BANE} \citep{yang2018binarized} & $\underset{\pm 0.015}{ 0.566}$ &$\underset{\pm 0.015}{ 0.743}$ &$\underset{\pm 0.019}{ 0.729}$&$\underset{\pm 0.011}{ 0.648}$&$\underset{\pm 0.009}{ 0.517}$\\[0.75em]
\textbf{TENE} \citep{yang2018enhanced} & $\underset{\pm 0.010}{ 0.658}$ &$\underset{\pm 0.011}{ 0.662}$ &$\underset{\pm 0.009}{ 0.775}$ &$\underset{\pm 0.016}{ 0.598}$&$\underset{\pm 0.022}{ 0.593}$\\[0.75em]
\textbf{FEATHER} \citep{feather} & $\underset{\pm 0.012}{ 0.649}$ &{$\underset{\pm 0.010}{ 0.805}$} &$\underset{\pm 0.015}{ 0.769}$&{ $\underset{\pm 0.039}{ 0.854}$}&$\underset{\pm 0.007}{ 0.539}$\\[0.75em]
\hline 
\textbf{2-Layer MLP} &$\underset{\pm 0.010}{ 0.706}$  &$\underset{\pm 0.011}{ 0.690}$ &$\underset{\pm 0.005}{ 0.783}$ &$\underset{\pm 0.010}{ 0.761}$ &$\underset{\pm 0.016}{ 0.565}$ \\[0.75em]
\textbf{Chebyshev} \citep{cnn_graph}& $\underset{\pm 0.005}{ 0.742}$ &$\underset{\pm 0.004}{ 0.855}$ &$\underset{\pm 0.005}{ 0.818}$ &$\underset{\pm 0.009}{ 0.838}$&$\underset{\pm 0.008}{ 0.564}$\\[0.75em]
\textbf{GCN} \citep{kipf2017semi}& {\color{red}$\mathbf{\underset{\pm 0.004}{ 0.767}}$ }&$\underset{\pm 0.003}{ 0.861}$ &$\underset{\pm 0.003}{ 0.822}$&$\underset{\pm 0.007}{ 0.854}$&$\underset{\pm 0.008}{ 0.545}$\\[0.75em]
\textbf{GAT} \citep{gat_iclr18}& $\underset{\pm 0.004}{ 0.748}$ & $\underset{\pm 0.010}{ 0.830}$& $\underset{\pm 0.002}{ 0.818}$& $\underset{\pm 0.008}{ 0.839}$&$\underset{\pm 0.009}{ 0.532}$\\[0.75em]
\textbf{SGConv} \citep{sgc_icml19}& $\underset{\pm 0.013}{ 0.699}$ & $\underset{\pm 0.005}{ 0.850}$& $\underset{\pm 0.010}{ 0.796}$&$\underset{\pm 0.005}{ 0.762}$&$\underset{\pm 0.006}{ 0.536}$\\[0.75em]
\textbf{ClusterGCN}  \citep{clustergcn_kdd19}& $\underset{\pm 0.006}{ 0.708}$ &$\underset{\pm 0.007}{ 0.836}$ &$\underset{\pm 0.005}{ 0.819}$&$\underset{\pm 0.010}{ 0.817}$&$\underset{\pm 0.005}{ 0.558}$ \\[0.75em]
\textbf{GraphSAGE} \citep{graphsage_nips17} & $\underset{\pm 0.008}{ 0.706}$ &$\underset{\pm 0.009}{ 0.840}$ &$\underset{\pm 0.007}{ 0.803}$ &$\underset{\pm 0.009}{ 0.846}$&$\underset{\pm 0.006}{ 0.554}$\\[0.75em]
\hline
\textbf{PDN} &$\underset{\pm 0.010}{ 0.764}$ &{\color{red}$\mathbf{\underset{\pm 0.008}{ 0.868}}$} &$\underset{\pm 0.004}{ 0.835}$ &{\color{red}$\mathbf{\underset{\pm 0.010}{ 0.875}}$}&{\color{red}$\mathbf{\underset{\pm 0.010}{ 0.584}}$}\\[0.75em]
\textbf{PDN EdgeConv}& $\underset{\pm 0.008}{ 0.711}$ &$\underset{\pm 0.008}{ 0.864}$ &$\underset{\pm 0.011}{ 0.833}$ &$\underset{\pm 0.009}{ 0.863}$&$\underset{\pm 0.007}{ 0.548}$\\[0.75em]
\textbf{PDN Multi-Scale} & $\underset{\pm 0.007}{ 0.740}$ & $\underset{\pm 0.009}{ 0.866}$& {\color{red}$\mathbf{\underset{\pm 0.013}{ 0.836}}$}&$\underset{\pm 0.009}{ 0.793}$&$\underset{\pm 0.009}{ 0.568}$\\ [0.75em]
\hline
\end{tabular}

}
\end{table}


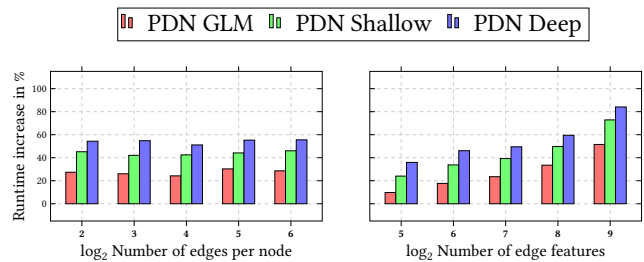
\begin{figure}[h!]
    \vspace{-2mm}
	\centering
	\begin{tikzpicture}[scale=0.45,transform shape]
	\tikzset{font={\fontsize{14pt}{12}\selectfont}}
	\begin{groupplot}[group style={group size=2 by 1,
		horizontal sep=40pt, vertical sep=50pt,ylabels at=edge left},
	width=0.54\textwidth,
	height=0.3375\textwidth,
	ymin=0,
	ymax=100,
	legend columns=3,
every tick label/.append style={font=\bf},
    y tick label style={
        /pgf/number format/.cd,
            fixed,
            fixed zerofill,
            precision=0,
        /tikz/.cd
    },
 enlarge x limits=true,
	grid=major,
	grid style={dashed, gray!40},
	scaled ticks=false,
	inner axis line style={-stealth}]










 \nextgroupplot[
   xlabel=$\log_2$ Number of edges per node,
    ybar=0pt,
      every tick/.style={
        black,
        semithick,
      },
    bar width=9pt,
    enlargelimits=0.15,
    ylabel={Runtime increase in \%},
    ytick={0,20,40,60,80,100},
    legend style={at={(0.5,-0.15)},
      anchor=north,legend columns=-1},
    symbolic x coords={2, 3, 4, 5, 6},
    xtick={2, 3, 4, 5, 6}]

\addplot [fill=red!55]  coordinates {
(2,27.374)
(3,26.038)
(4,24.225)
(5,30.276)
(6,28.608)
};
\addplot [fill=green!55]  coordinates {
(2,45.182)
(3,42.058)
(4,42.472)
(5,44.161)
(6,46.009)
};
\addplot [fill=blue!55] coordinates {
(2,54.326)
(3,54.76)
(4,51.096)
(5,55.266)
(6,55.516)
};



 \nextgroupplot[
   xlabel=$\log_2$ Number of edge features,
    ybar=0pt,
      every tick/.style={
        black,
        semithick,
      },
    bar width=9pt,
    enlargelimits=0.15,
    legend columns=3,
    legend image post style={solid},
    legend style={at={(0.5,-0.25)},nodes={scale=1.5, transform shape}, 
      anchor=north,legend columns=-1},
yticklabels={,,,,,},
ytick={0,20,40,60,80,100},
    symbolic x coords={5, 6, 7, 8, 9},
    xtick={5, 6, 7, 8, 9},
    	legend style = { column sep = 10pt, legend columns = 1, legend to name = grouplegend}   ]

\addplot [fill=red!55]  coordinates {
(5,9.778)
(6,17.688)
(7,23.487)
(8,33.516)
(9,51.527)};
\addlegendentry{PDN GLM}
\addplot [fill=green!55]  coordinates {
(5,24.023)
(6,33.737)
(7,39.255)
(8,49.778)
(9,72.853)};
\addlegendentry{PDN Shallow}
\addplot [fill=blue!55] coordinates {
(5,35.916)
(6,46.083)
(7,49.474)
(8,59.484)
(9,84.09)};\addlegendentry{PDN Deep}
	\end{groupplot}

	\node at ($(group c2r1) + (-4.5cm,3.6cm)$) {\ref{grouplegend}}; 
	\end{tikzpicture}
    \vspace{-2mm}	
	\caption{The relative runtime increase (compared to spectral graph convolutions) needed for training PDN models on synthetic datasets.
	}\label{fig:runtime}
    \vspace{-3mm}	
\end{figure}

\subsection{Relative runtime}
The time complexity of training a traditional spectral graph convolutional networks is $\mathcal{O}(|E| F)$ while a Pathfinder Discovery Network has a time complexity of $\mathcal{O}(|E| (F+D))$. Using synthetic data, we compare the relative runtime of PDNs in a number of scenarios to provide a better empirical understanding of what the additional time complexity means in practice. Experimental details are summarized in Appendix \ref{app:runtime}.

\textbf{Findings and Discussion.} The relative runtime is shown in Figure \ref{fig:runtime}. The results are in line with the runtime complexities discussed above: increasing the number of edges does not increase the relative runtime of the PDNs, but increasing the edge feature count does increase the relative runtime. We also see that more complex (deeper) edge aggregation models are slower.

\subsection{Edge feature importance}

Model interpretability is an important part of developing deep neural networks. Architectures with interpretable weights can provide novel insights on the structure of data, while also making validation, inspection, and debugging significantly easier. We believe that PDNs can add significant interpretability to graph learning tasks when we frame the learned weights as an attention mechanism over the input graphs. In this set of experiments we discuss two scenarios when learned PDN weights have direct interpretations. 
 \input{./figures/attention_evolution.tex}

\textbf{Attention on proximity.} 
In this experiment, we use the multi-scale model described in Section \ref{subsec:multiscale}. We utilize the first 5 normalized adjacency matrix powers as input similarity graphs and apply the hyperparameters described in Section \ref{subsec:performance}. We train this model on 100-shot learning tasks, and report the mean weight for each adjacency power from 100 repetitions (see Figure \ref{att:evolution}).

Based on these learned weights, we observe that the model has learned to prioritize messages that come from the first order neighbourhoods of vertices -- in other words, the PDN attends to closer neighbors more. We note that the Cora and Pubmed graphs exhibit high homophily between nodes, suggesting that the model's weighting scheme is well motivated. Interestingly, we also observe that the importance of information coming from the second hop starts to decline between 50-100 epochs; not coincidentally, this is around when peak test accuracy is reached, after which we observe a decline (test accuracy not shown). This implies that graph neural network models overfit to information coming from the first order proximity of individual data points. More importantly, the added interpretability from PDN allows us to observe exactly where the overfitting is occurring.

\begin{figure}[t!]
\begin{tikzpicture}[scale=0.750]
\begin{axis}[
table/col sep=comma,
xbar=0pt, xmin=0,xmax=0.16,
xlabel=Attention,
xtick={0.0,0.04,0.08,0.12,0.16},
yticklabels from table={reshaped.csv}{Region},
yticklabel style={text height=1.2ex},
ytick=data,
width=0.5\textwidth,
y=0.9cm,
enlarge y limits={abs=0.5},
bar width=3pt,
/pgf/number format/fixed,
axis lines*=left,
xmajorgrids=true,
legend columns=5,
legend style={at={(0.5,1.15)},anchor=north},
legend entries={Deezer, Facebook, Pubmed, Cora, Citeseer},
reverse legend,
area legend,
]
\addplot [fill=green!50] table [y expr=-\coordindex, x=Citeseer] {reshaped.csv};

\addplot [fill=red!50] table [y expr=-\coordindex, x=Cora] {reshaped.csv};

\addplot [fill=cyan!50] table [y expr=-\coordindex, x=Pubmed] {reshaped.csv};

\addplot [fill=orange!50] table [y expr=-\coordindex, x=Facebook] {reshaped.csv};

\addplot [fill=blue!50] table [y expr=-\coordindex, x=Deezer] {reshaped.csv};

\end{axis}
\end{tikzpicture}
  \caption{Comparison of average PDN attention scores (using a single pathfinder neuron) on edge similarity scores for the real world datasets.}
  \label{fig:attention_sim}
  \end{figure}
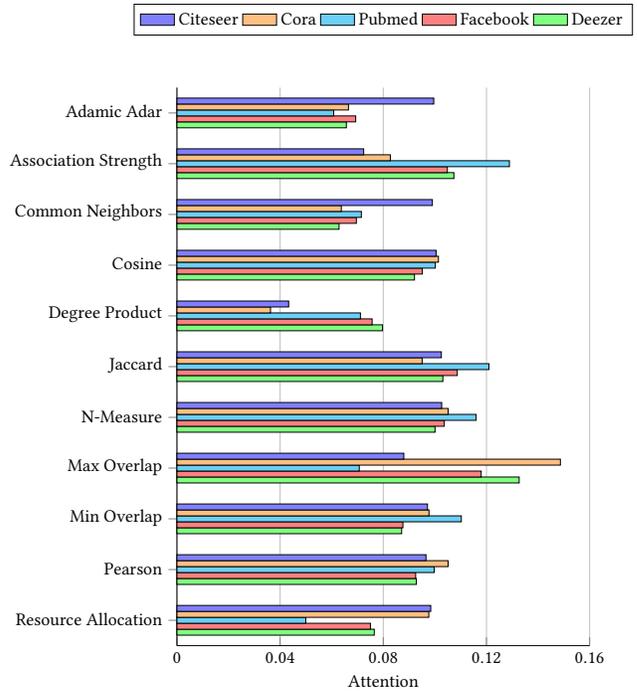

\textbf{Attention on neighbourhood similarity.}
Using the similarity scores listed in Table \ref{tab:edge_features} of Appendix \ref{app:edge_features} we train a Linear PDN model with the hyperparameter settings described in Appendix \ref{app:real_world_settings}. As a reminder, this implementation of the pathfinder layer uses softmax activations and does not have a hidden layer -- in this setting, the weights can be interpreted as attention. For the citation graph and social network datasets we plot the average attention score (calculated from 10 training runs) for a selected subset of edge scores in Figure \ref{fig:attention_sim}.

The results show that unnormalized edge similarity scores such as the \textit{degree product} and \textit{common neighbours} tend to receive low attention when the edge weight aggregation happens. On most datasets similarity metrics which are normalized and do not consider the degree of shared neighbors (e.g. \textit{association strength} and \textit{minimal overlap}) receive relatively high attention.

\section{Conclusion}\label{sec:conclusions}
In this paper we proposed pathfinder discovery networks (PDN), a graph neural network architecture for learning a message passing graph from a multiplex graph defined on a fixed set of nodes. 
Our modular architecture allows for joint training of a graph neural network and the pathfinder discovery layer, which in turn allows practitioners to find the optimal message passing graph for a specific supervised task. We examine the comparative characteristics of PDNs, concluding that PDNs are significantly more expressive and resilient than existing approaches. We then describe general extensions of our model, which allow for the definition of multi-scale graph convolutional layers and edge convolutions without edge features.

In our empirical analysis, we establish that PDNs have competitive predictive performance on various node classification tasks. We showed that the relative runtime increase of PDNs is independent of the dataset size in terms of edge set cardinality. And finally, we examined the weights of the graph aggregation model from the lens of learned attention. 

We believe there are many exciting areas of future work. We are particularly excited about the possibility of extracting the PDN-learned graph for use in other tasks. We intuit that it would be possible to learn several PDN graphs for many different kinds of supervised tasks, and then combine \textit{those} graphs in another PDN. We believe that this yet-unexplored use case will be very important in improving abstract notions of `graph accuracy' for a wide range of datasets, while simultaneously opening new areas of transfer learning on graphs.

\bibliographystyle{ACM-Reference-Format}

\bibliography{bibliography}

\appendix

\section{Synthetic node classification experimental settings}\label{app:synthetic_settings}
The default setting for synthetic graph generation are as follows: we generate graphs with $C=3$ label classes and $n=500$ nodes per class; we set edge probabilities to $P=0.01$ and $Q=0.005$; we set feature dimensions to $F=32$ and $D=32$; and we set feature correlations to $\sigma_F=5.0$ and $\sigma_D=2.0$ standard deviations. We modulate these hyperparameters in the experimental scenarios described in Table \ref{tab:scenarios}, where each scenario modifies a single parameter from the defaults described above. For each scenario, we used 80\%/20\% train-test splits, and report the average of 10 synthetic graph generation\%/model training cycles. 

For our synthetic experiments, we use GCN, GAT, and DeepWalk as baselines, with the hyperparameter settings described in \cite{mixhop_icml19}. To this set, we add the following baselines and corresponding hyperparameters: 

\begin{itemize}
\item \textit{AAPNP} \citep{ppnp_iclr19, bojchevski2020pprgo}: The feedforward component of the model has 32 filters and we did 10 personalized pagerank approximation iterations with a teleport probability of 0.2.
\item \textit{SGCONV} \citep{sgc_icml19}: We used information from the $2^{nd}$ order proximity of the normalized adjacency matrix with 32 dimensional filters.
\item \textit{ClusterGCN} \citep{clustergcn_kdd19}: We used the settings of the \textit{Spectral GCN} model on the graph pre-clustered by the METIS community detection algorithm \cite{metis}.
\end{itemize} 

By comparison, we construct a PDN with a single hidden layer containing 16 pathfinder neurons and a ReLU activation function \cite{nair2010rectified} in the hidden layer, followed by a softmax activation function in the output layer of the pathfinder module. On top of the pathfinder layers, we add a standard 2-hop spectral GCN \cite{kipf2017semi} with a hidden layer dimension size of 32.  All models, including PDNs, were trained using Adam \citep{adam} with a learning rate of $10^{-2}$, over $200$ training epochs. Where relevant, we used a dropout value of 0.5 and an $l_2$ weight regularization coefficient of $10^{-3}$. All models were implemented using the PyTorch Geometric framework \cite{pytorch_geometric}.

\section{Multiplex benchmark dataset descriptive statistics}\label{app:descriptives_multiplex}
We used publicly available multiplex attributed webgraph datasets for the binary node classification experiments \cite{dmgi}. We summarized the descriptive statistics of the graph layers in Table \ref{tab:multiplex_descriptives}. We would like to point out that the layer wise characteristics of the networks are remarkably different for these two datasets.
\begin{table}[h!]
\caption{Descriptive statistics of the multiplex webgraphs (individual layers) used for node classification performance evaluation and comparison in our work.}\label{tab:multiplex_descriptives}
	\centering{\footnotesize
\begin{tabular}{lcccccc}
\textbf{Dataset} &\textbf{Layers}& \textbf{Nodes} & \specialcell{\textbf{Density}} & \specialcell{\textbf{Clustering}\\\textbf{Coefficient}} & \specialcell{\textbf{Unique}\\\textbf{Features}}&\textbf{Classes}\\
\hline
\textbf{IMDB} &2& 3550 & \specialcell{0.005\\0.001} & \specialcell{0.509\\1.000} &2000&2\\
\hline
\textbf{AMC} &2& 3025 & \specialcell{0.242\\0.004} & \specialcell{1.000\\0.687} &1870&2\\
\hline
\end{tabular}
}
\end{table}

\section{Multiplex node classification experimental settings} \label{app:multiplex_settings}
We created 10 seeded 100-shot learning splits for evaluation, because of this the mean performance metrics are comparable across models as there is no variation coming from the splits. The PDN had a single hidden layer with 2 neurons, the other hyperparameters were the same as the ones described in Appendix \ref{app:real_world_settings}. One of the supervised baselines was a spectral GCN \cite{kipf2017semi} which used the union of edge sets from the graph layers, this model also used the experimental settings from \ref{app:real_world_settings}. The other supervised reference models Multi-GCN \cite{khan2019multi} and DMGI \cite{dmgi} used the default hyperparameters from the experimental section of the respective research papers. The evaluation of the unsupervised techniques MVN2Vec \cite{shi2018mvn2vec}, MELL \cite{matsuno2018mell} and MNE \cite{mne} used a two stage upstream and downstream learning setup. First, we trained embeddings with hyperparameters from the original papers. Second, we trained a scikit-learn  \cite{scikit}  logistic regression on the embedding features using the default settings.

\section{Dataset descriptive statistics}\label{app:descriptives}

\begin{table}[h!]
\caption{Descriptive statistics of the attributed citations graphs and social networks used for node classification performance evaluation and comparison in our work.}\label{tab:desriptives}

	\centering{\footnotesize
\begin{tabular}{lrcccc}
\textbf{Dataset} & \textbf{Nodes} & \specialcell{\textbf{Clustering}\\\textbf{Coefficient}} & \textbf{Density} & \specialcell{\textbf{Unique}\\\textbf{Features}}&\textbf{Classes}\\
\hline
Cora & 2,708&0.094&0.002&1,432&7\\[0.25em]
Citeseer&3,327&0.130& 0.001&3.703&6\\[0.25em]
Pubmed  & 19,717&0.054&0.001&500&3\\[0.25em]
\hline
Facebook Page-Page&22,470     &0.232&0.001&4,714&4\\[0.25em]
Deezer Europe&28,281&0.096&0.001&31,240&2\\[0.25em]
\hline
\end{tabular}
}
\vspace{-3mm}
\end{table}

\section{Real world node classification experimental settings}\label{app:real_world_settings}

We evaluated proximity preserving node embedding techniques \cite{tang2015line, deepwalk, node2vec, hope,netmf,diff2vec, gemsec}, including multi-scale methods  \citep{perozzi2017don, grarep}. We also included a range of attributed node embedding methods \citep{yang2015network, yang2018enhanced} and attributed methods that incorporate node attribute information from multiple hops \citep{yang2018binarized}. Each of the upstream node embeddings was trained with the default hyperparameter settings of the Karate Club package \cite{karateclub} -- 128 dimensional node embeddings which have a comparable number of free parameters. The downstream model was an $l_1$ regularized multinomial logistic regression (softmax) classifier pulled from scikit-learn \cite{scikit}.

For supervised baselines, we used GNN hyperparameter settings, training setup, and citation graph results from \citep{mixhop_icml19}, specifically the performance of the two layer feedforward neural network, \textit{Chebyshev GCN} \citep{cnn_graph}, \textit{Spectral GCN} \citep{kipf2017semi} and  \textit{GAT} \citep{gat_iclr18}. For comparison, we examine three PDNs: the basic Pathfinder Discovery Network, the PDN EdgeConv method described in \ref{subsec:edgeconv}, and the PDN Multi-Scale method described in \ref{subsec:multiscale}. The default PDN has a single hidden layer with 16 pathfinder neurons and uses the neighbourhood similarity metrics \cite{adamic2003friends, egghe2009relation, zhou2009predicting, opsahl2010node} described in Table \ref{tab:edge_features} of Appendix \ref{app:edge_features} as input features. In addition, our edge convolutional model uses information from the $1^{st}$ and $2^{nd}$ hop, while the cheap multi-scale model uses information up to the $2^{nd}$ order proximity. We use the same hyperparameters and optimizer settings discussed in Appendix \ref{app:synthetic_settings}. All models were trained on a 100-shot learning experiment where we calculated the average node classification accuracy on the test set based on 10 seeded train-test splits.

\section{Relative runtime evaluation experimental settings}\label{app:runtime}
We generate Watts-Strogatz graphs \cite{watts} with $10^{12}$ nodes, $2^{4}$ edges per node and a rewiring probability of 0.5. In addition, we sample $F=2^7$ node and $D=2^7$ edge features using Gaussians and draw labels for the nodes from $C=4$ classes uniformly. We calculate the average epoch runtime for a spectral GCN \cite{kipf2017semi}, a generalized linear PDN, a shallow PDN with $\{32\}$ neurons, and a deep PDN with $\left \{32, 16\right\}$ neurons in the hidden layers.

\section{Tie strength edge features}\label{app:edge_features}

\begin{table}[h!]
\caption{Tie strength scoring functions for edge $(u,v) \in E$ used as edge features of the Pathfinder Discovery Networks.}\label{tab:edge_features}
{\small
\begin{tabular}{c c}
\hline
\textbf{Name} & \textbf{Definition}  \\ \hline
Adamic-Adar            &   $\sum\limits_{w \in N(u) \cap N(v)} \frac{1}{\log{|N(w)|}}$        \\[0.75em]
Association Strength   &      $\frac{|N(u) \cap N(v)|}{|N(u)| \cdot |N(v)|}$    \\[0.75em]
Common Neighbors       &    $|N(u)\cap N(v)|$     \\[0.75em]
Cosine                 &   $\frac{|N(u) \cap N(v)|}{\sqrt{|N(u)| \cdot |N(v)|}}$       \\[0.75em]
Degree Product         &    $|N(u)|\cdot |N(v)|$        \\[0.75em]
Jaccard                &  $\frac{|N(u) \cap N(v)|}{|N(u) \cup N(v)|}$           \\[0.75em]
Max Overlap            &  $\frac{\max(|N(u)| , |N(v)|)}{|N(u) \cap N(v)|}$             \\[0.75em]
Min Overlap            &  $\frac{\min(|N(u)| , |N(v)|)}{|N(u) \cap N(v)|}$           \\[0.75em]
N-Measure              &$ \frac{\sqrt{2} |N(u) \cap N(v)|}{\sqrt{|N(u)|^2 +|N(v)|^2 }}$  \\[0.75em]
Pearson Correlation    & $\frac{|V|\cdot |N(u)\cap N(v)|-|N(u)|\cdot |N(v)|}{\sqrt{|V|\cdot |N(u)|-|N(u)|^2}\cdot\sqrt{|V|\cdot |N(v)|-|N(v)|^2}}
  $           \\[0.75em]
Resource Allocation    &    $\sum\limits_{w \in N(u) \cap N(v)} \frac{1}{|N(w)|}$         \\
 \hline
\end{tabular}
}
\end{table}

\end{document}